\def\paperlanguage{}
\pdfoutput=1 % for arxiv

\documentclass[letterpaper, 10 pt, conference]{ieeeconf}  % Comment this line out if you need a4paper

\usepackage{bm}
\usepackage{cite}
\usepackage{flushend}
\include{preamble}

\IEEEoverridecommandlockouts                              % This command is only needed if
\title{\LARGE \textbf
  {
    \switchlanguage%
    {%
      Learning of Balance Controller Considering Changes in Body State\\for Musculoskeletal Humanoids
    }%
    {%
      時間的身体変化を考慮した筋骨格ヒューマノイドのバランス制御学習
    }%
  }
}

\author{Kento Kawaharazuka$^{1}$, Yoshimoto Ribayashi$^{1}$, Akihiro Miki$^{1}$, Yasunori Toshimitsu$^{1}$,\\Temma Suzuki$^{1}$, Kei Okada$^{1}$, and Masayuki Inaba$^{1}$% <-this % stops a space
  \thanks{$^{1}$ The authors are with the Department of Mechano-Informatics, Graduate School of Information Science and Technology, The University of Tokyo, 7-3-1 Hongo, Bunkyo-ku, Tokyo, 113-8656, Japan.
    {\texttt\small [kawaharazuka, ribayashi, miki, toshimitsu, t-suzuki, k-okada, inaba]@jsk.t.u-tokyo.ac.jp}
  }
}
\begin{document}

\maketitle
\thispagestyle{empty}
\pagestyle{empty}

%%%%%%%%%%%%%%%%%%%%%%%%%%%%%%%%%%%%%%%%%%%%%%%%%%%%%%%%%%%%%%%%%%%%%%%%%%%%%%%%
\begin{abstract}
  \switchlanguage%
  {%
    The musculoskeletal humanoid is difficult to modelize due to the flexibility and redundancy of its body, whose state can change over time, and so balance control of its legs is challenging.
    There are some cases where ordinary PID controls may cause instability.
    In this study, to solve these problems, we propose a method of learning a correlation model among the joint angle, muscle tension, and muscle length of the ankle and the zero moment point to perform balance control.
    In addition, information on the changing body state is embedded in the model using parametric bias, and the model estimates and adapts to the current body state by learning this information online.
    This makes it possible to adapt to changes in upper body posture that are not directly taken into account in the model, since it is difficult to learn the complete dynamics of the whole body considering the amount of data and computation.
    The model can also adapt to changes in body state, such as the change in footwear and change in the joint origin due to recalibration.
    The effectiveness of this method is verified by a simulation and by using an actual musculoskeletal humanoid, Musashi.
  }%
  {%
    筋骨格ヒューマノイドはその柔軟性と冗長性によるモデル化困難な身体からバランス制御が難しく, その身体状態は時間的に変化することがある.
    通常のPID制御等では逆にバランスが不安定になってしまうような場合も存在する.
    そこで本研究では, 下腿から足にかけての関節角度や筋張力, 筋長, ZMPの相関モデルを学習し, これを使ってバランス制御を実行する手法を提案する.
    また, 変化する身体状態の情報をParametric Biasによりモデル内に埋め込み, これをオンラインで学習することで現在の身体状態を推定し適応する.
    これにより, データ量や計算量の観点から本モデルで完全な全身のダイナミクスを学習することは難しく, モデルに直接考慮されない上半身姿勢等の身体変化に適応することが可能となる.
    また, 履いている靴の変化や再キャリブレーションによる関節原点の変化等の身体特性変化にも適応することが可能となる.
    本手法の有効性を筋骨格ヒューマノイドMusashiのシミュレーションと実機において検証した.
  }%
\end{abstract}

\section{INTRODUCTION}\label{sec:introduction}
\switchlanguage%
{%
  A variety of musculoskeletal humanoids have been developed so far \cite{gravato2010ecce1, asano2016kengoro, kawaharazuka2019musashi}.
  Due to the flexibility and redundancy of their bodies, all of them are very difficult to control in the same way as ordinary axis-driven robots.
  Various learning-based control methods have been proposed for them.
  In \cite{nakanishi2005pedaling}, a pedaling operation is acquired by self-repetitive learning.
  \cite{diamond2014reaching} proposes to control the upper body reaching motion using reinforcement learning.
  In \cite{marjaninejad2019tendon}, for a relatively simple system with one or two joints, the relationship among joints, muscles, and tasks is trained, and a robot is controlled using the trained neural network.
  In \cite{kawaharazuka2018online, kawaharazuka2020autoencoder}, for a more complex system, the relationship among joint angle, muscle tension, and muscle length is modelized by a neural network, which is trained and applied mainly to upper body control and state estimation.
  \cite{kawaharazuka2020dynamics} has succeeded in recognizing grasped objects and stabilizing tool grasping by learning the dynamics of a musculoskeletal hand.
  These methods have made it possible for complex musculoskeletal robots to acquire the ability to control themselves autonomously.

  On the other hand, the balance control of these robots is still difficult.
  In the case of balance control, data collection itself is difficult, because data must be acquired while the robot is in a balanced state.
  Therefore, none of the studies described so far deals with balancing.
  Although a simple balance control using PID has been implemented \cite{asano2019anklehip}, it is difficult to say that its balance has improved because the convergence of zero moment point has not been evaluated, and the success rate of the step-out experiment is extremely low.
  In addition, there is usually a strong human parameterization according to the structure of the robot, and the robot does not acquire the balance control autonomously.
  Exempting musculoskeletal humanoids, methods to solve this problem have been developed by using real2sim \cite{hwangbo2019anymal} and sim2real \cite{openai2019solving} to perform reinforcement learning in simulation environments.
  A safe learning with dynamics balancing models \cite{ahn2020locomotion} and locomotion generation with learning by cheating \cite{miki2022anymal} has also been developed.
  In addition, for quadruped robots, where balance is relatively easy to handle, running motion is generated only by learning on the actual robot \cite{yang2019legged}.
  On the other hand, it is very difficult to construct a model of a complex body such as the musculoskeletal humanoid in a simulation, and also it is challenging for the actual bipedal robot to collect data for model learning while maintaining balance.
  Even if we can construct a simulation, it is difficult to transfer the simulation model to the actual robot because of the large differences in muscle elongation, friction, muscle paths, etc.
  Therefore, it is desirable to acquire balance control autonomously by learning the relationships among various sensor values only in the actual robot.
  Also, it is necessary to solve the problem of the difficult data collection in the actual robot.
}%
{%
  これまで様々な筋骨格ヒューマノイドが開発されてきた \cite{gravato2010ecce1, asano2016kengoro, kawaharazuka2019musashi}.
  そのどれもが身体の柔軟性と冗長性から非常に制御が難しく, 通常の軸駆動型ロボットと同じような制御を行うことは困難である.
  これらに対して様々な学習型制御手法が提案されてきた.
  \cite{nakanishi2005pedaling}では自己繰り返し学習によるペダリング操作の獲得を行っている.
  \cite{diamond2014reaching}では強化学習を使った上半身のリーチング動作制御を提案している.
  \cite{marjaninejad2019tendon}では, 1関節や2関節の小さな系において, 関節や筋, タスクの関係をニューラルネットワークにより学習し制御している.
  \cite{kawaharazuka2018online, kawaharazuka2020autoencoder}ではより複雑な系自由度を持つ系について, 関節角度-筋張力-筋長の関係をニューラルネットワークでモデル化し, これを学習して制御や状態推定等に応用している.
  \cite{kawaharazuka2020dynamics}では筋骨格ハンドのダイナミクスを学習することで把持物体認識や道具把持安定化に成功している.
  これらにより, 複雑なロボットが自律的に自身を制御する能力を獲得できるようになってきた.

  一方, これらのロボットのバランス制御は未だに難しい.
  バランス制御の場合, 常にバランスを保った状態でデータを取得しなければならないため, そもそもデータ取得自体が難しい.
  そのため, これまでに説明した研究のどれもがバランスについて扱っていない.
  PID等による簡単なバランス制御は行われているが\cite{asano2019anklehip}, zmpの収束を評価しておらず, 踏み出し実験の成功率も33\%と, バランス性能が向上したとは言い難い.
  また, 大抵ロボットや構造に応じた人間の強いパラメタライズが入っており, ロボットが自律的にバランス制御を獲得していない.
  筋骨格ヒューマノイドを除けば, これを解決する手法としては, Real2Sim \cite{hwangbo2019anymal}, Sim2Real \cite{openai2019solving}を用いた強化学習手法が開発されている.
  dynamics modelを用いたsafe learningを併用した手法\cite{ahn2020locomotion}やlearning by cheatingを用いたlocomotion学習手法\cite{miki2022anymal}も開発されている.
  また, 4脚ロボットのようなバランスが扱いやすい系については, 実機のみにおける走行の研究も存在する\cite{yang2019legged}.
  一方筋骨格ヒューマノイドのような複雑な身体はシミュレーションにおいてモデルを構築すること自体が非常に難しく, 2脚であるためバランスを取りながらデータ収集が難しい.
  シミュレーションを構築できても, 筋の伸びや摩擦, 筋経路等の違いが大きく, 実機に転移することは難しい.
  そのため, 実機だけで様々なセンサ値の関係性を学習することでバランス制御を自律的に獲得していくことが望ましい.
  また, 実機におけるデータ収集の問題点を解決する必要がある.
}%

\begin{figure}[t]
  \centering
  \includegraphics[width=0.9\columnwidth]{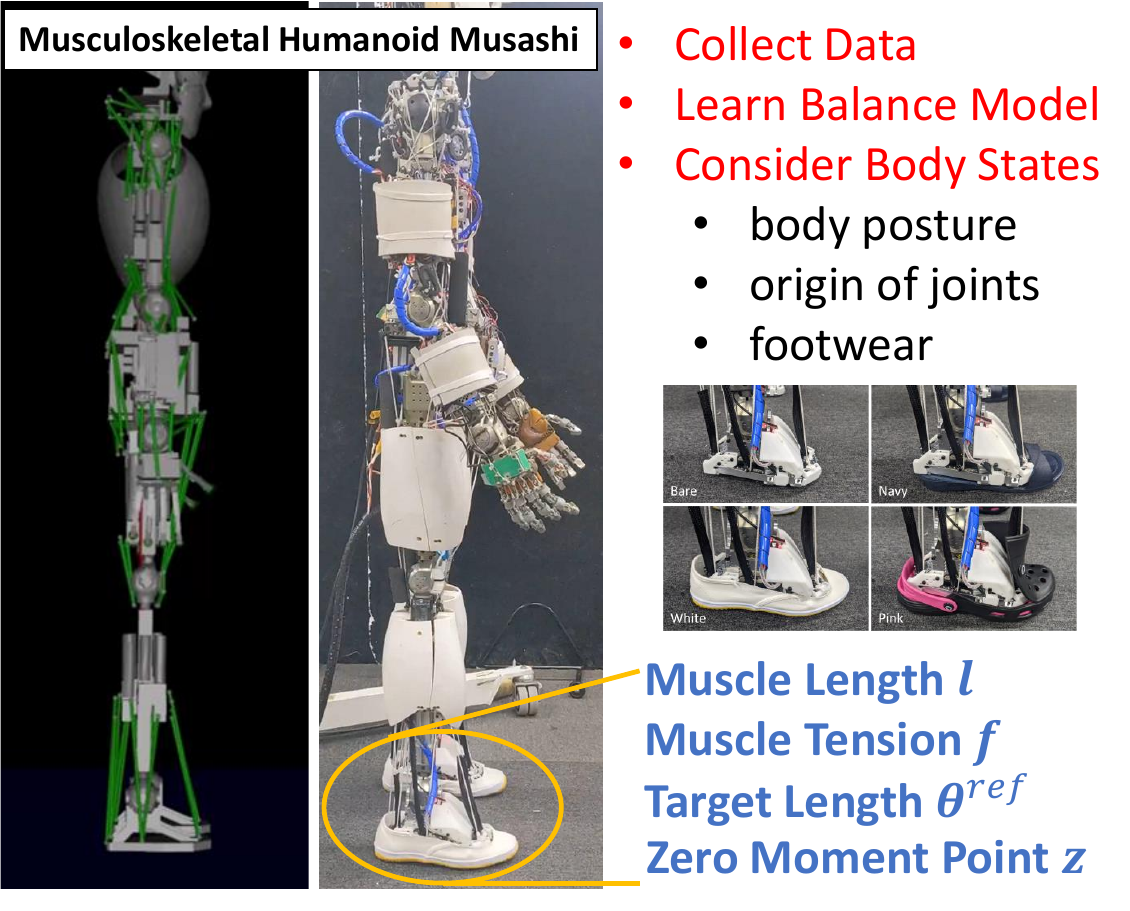}
  \vspace{-1.0ex}
  \caption{The concept of this study.}
  \label{figure:concept}
  \vspace{-3.0ex}
\end{figure}

\begin{figure*}[t]
  \centering
  \includegraphics[width=1.8\columnwidth]{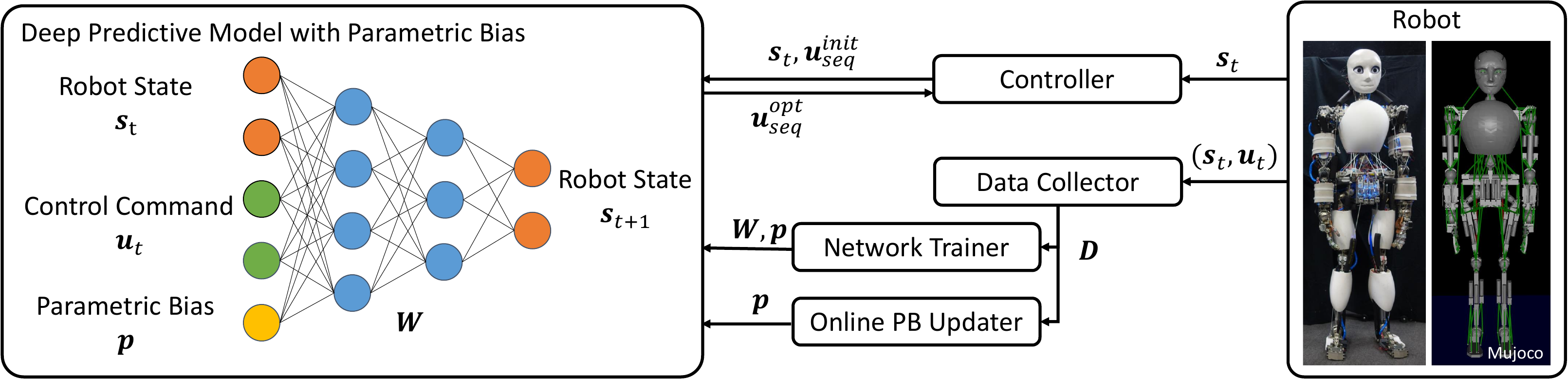}
  \vspace{-1.0ex}
  \caption{The overall system of balance control for musculoskeletal humanoids using a deep predictive model with parametric bias.}
  \label{figure:whole-system}
  \vspace{-3.0ex}
\end{figure*}

\switchlanguage%
{%
  In addition, there is a problem that there are many changes in dynamics of the body that cannot be directly represented in the model due to changes in the current body state.
  First, as it is difficult to learn the dynamics of all the sensor values of the whole body since it requires a huge amount of data, we will learn only some of the dynamics.
  In other words, if we learn the dynamics model among the ankle joints, muscles, and zero moment point, the changes in dynamics caused by the changes in upper body posture, etc., which are not included in the model, cannot be taken into account.
  Second, changes in the origin of muscles and joints due to irreproducible calibration, which is characteristic to musculoskeletal structures that usually do not have joint angle sensors, cannot be taken into account.
  In addition, if the shoes worn by the humanoid change, the dynamics will change significantly, which must also be taken into account.
  These changes in dynamics should be handled as disturbances or embedded as some low-dimensional parameters, and the balance control should be adapted to them.
  However, the former method that handles dynamics changes as disturbances usually requires accurate body models and prior knowledge of the distribution of disturbances.
  Also, since humans do not treat changes in the shoes that they are wearing as disturbances, but rather their walking style changes in response to the shoes, we believe that a new control method that incorporates this phenomenon is important.
  In this study, we apply the mechanism of parametric bias \cite{tani2002parametric} to this problem.
  Parametric bias is a bias parameter that allows multiple attractor dynamics to be implicitly embedded in a neural network, and has been mainly used in the context of imitation learning.
  In this study, we use this variable to learn a predictive model of various sensors for body balance, and embed the information of the changes in body state described so far in parametric bias.
  By learning this predictive model and estimating the changes in body state online, it is possible to perform balance control while adapting to the current body state (\figref{figure:concept}).
  Note that, although parametric bias is also used in \cite{kawaharazuka2020dynamics, kawaharazuka2022cloth}, this study examines data collection methods and application to changes in body state including wearing shoes for balance control of a flexible body.

  The purpose of this study is to develop a balance control system for humanoids with complex and flexible bodies that are difficult to modelize and whose body states change over time.
  Therefore, we develop a balance control system for musculoskeletal humanoids using a Deep Predictive Model with Parametric Bias (DPMPB).
  This enables not only autonomous learning of balance control but also adaptive control to changes in upper body posture and shoes, which are not directly included in the model.
  The contributions of this study are summarized as follows.
  \begin{itemize}
    \item Data collection for balance control in the actual musculoskeletal humanoid
    \item Embedding of changes in body state including wearing shoes into the model using parametric bias
    \item Online adaptation to the current body state and balance control using DPMPB
  \end{itemize}
  This method is applied to a simulation and an actual musculoskeletal humanoid, Musashi \cite{kawaharazuka2019musashi}, to confirm its effectiveness.
}%
{%
  加えて, 現在の身体状態変化に起因する, 直接モデル内に表現できない身体のダイナミクス変化が多く存在するという問題がある.
  まず, 全身のセンサ値に関するダイナミクスを学習するには膨大なデータが必要なためこれは困難であり, 一部のダイナミクスのみを学習する必要がある.
  つまり, 足首の関節と筋, ZMPに関するダイナミクスを学習した場合, ここに含まれない上半身姿勢等の変化によるダイナミクス変化は考慮されない.
  また,一般的に関節角度センサを持たない筋骨格構造に特有な, 再現性の低いキャリブレーションによる筋原点や関節原点の変化が考慮されていない.
  この他にも, 履いている靴が変化したらダイナミクスは大きく変化し, これらも考慮する必要がある.
  これらダイナミクス変化は, 外乱として扱う, または何らかの低次元パラメータとして埋め込み, これらに適応した制御を行う必要がある.
  ここで, ダイナミクス変化を外乱として扱うロバストな制御手法は正確な身体モデルや受ける外乱の分布を事前に知っておく必要があり, 本研究には適さない.
  また, 人間も履いている靴の変化を外乱として扱うわけではなく, 靴に応じて歩き方が変わるため, その現象を取り込んだ新しい制御法が必要であると考える.
  本研究ではこの問題点に対して, Parametric Bias \cite{tani2002parametric}の仕組みを応用する.
  Parametric Biasは複数のアトラクターダイナミクスを暗黙的に埋め込むことが可能な仕組みで, これまで主に模倣学習の文脈で利用されてきた\cite{ogata2005extracting, kawaharazuka2021imitation}.
  本研究ではこれを, 身体バランスに関する様々なセンサ間の予測モデルを学習する際に用いることで, Parametric Biasにこれまで説明した身体状態変化の情報を埋め込む.
  これをオンラインで学習・推定することで, 現在の身体状態に適応しながらバランス制御を実行することが可能になる.
  なお, \cite{kawaharazuka2020dynamics, kawaharazuka2022cloth}でも同様にparametric biasを利用しているが, 本研究は様々な靴におけるバランスということで, データ収集, これらバランス制御や靴変化への応用について検証するものである.

  本研究では, モデル化困難な身体を持ち, かつその身体状態が変化する複雑で柔軟なヒューマノイドのバランス制御を開発することを目的とする.
  これにより, 自律的な学習によりバランス制御を構築するだけでなく, 直接モデル内に含まれない上半身姿勢の変化や履いている靴の変化にまで適応した制御を可能とする.
  そこで, 筋骨格ヒューマノイドにおけるParametric Biasを含む深層予測モデル(Deep Predictive Model with Parametric Bias, DPMPB)を使ったバランス制御を行う.
  本研究のコントリビューションを以下にまとめる.
  \begin{itemize}
    \item バランス制御のためのデータ収集
    \item Parametric Biasを使ったモデルに対する靴変化を含む身体状態変化情報の埋め込み
    \item DPMPBを使った現在の身体状態への逐次適応とバランス制御
  \end{itemize}
  筋骨格ヒューマノイドMusashi \cite{kawaharazuka2019musashi}のシミュレーションと実機に本手法を適用し, その有効性を確認する.
}%

\section{Balance Control of Musculoskeletal Humanoids Using Deep Predictive Model with Parametric Bias} \label{sec:proposed}
\switchlanguage%
{%
  The overall system of balance control using DPMPB is shown in \figref{figure:whole-system}.
}%
{%
  本研究におけるDPMPBを中心とした全体システムを\figref{figure:whole-system}に示す.
}%

\subsection{Network Structure of DPMPB}
\switchlanguage%
{%
  The network structure of DPMPB is shown below,
  \begin{align}
    \bm{s}_{t+1} = \bm{h}(\bm{s}_{t}, \bm{u}_{t}, \bm{p}) \label{eq:dpnpb}
  \end{align}
  where $t$ is the current time step, $\bm{s}$ is the sensor state, $\bm{u}$ is the control input, $\bm{p}$ is parametric bias, and $\bm{h}$ is the network of DPMPB.
  In this study, for the balance control in the musculoskeletal humanoid, we directly deal with the state of joints and muscles related to the ankles, while the posture of the upper body is implicitly handled by parametric bias.
  Therefore, we set $\bm{s}=\begin{pmatrix}\bm{z}^{T} & \bm{f}^{T} & \bm{l}^{T}\end{pmatrix}^{T}$ and $\bm{u}=\theta^{ref}$.
  Here, $\bm{z}$ is zero moment point (ZMP), $\{\bm{f}, \bm{l}\}$ is \{muscle tension, muscle length\} regarding the ankles of both legs, and $\theta^{ref}$ is the target joint angle of the ankles.
  Note that $\bm{z}$ is 2-dimensional ($z_x$ for $x$-direction and $z_y$ for $y$-direction), and the dimension of $\{\bm{f}, \bm{l}\}$ depends on the robot configuration.
  Although $\theta^{ref}$ can have roll and pitch angles for both legs, we assume the angles for both legs to be the same and $\theta^{ref}$ to be 1-dimensional only for the pitch axis in this study, for simplicity.
  Parametric bias $\bm{p}$ is an input variable that can embed implicit differences in dynamics, and in this study, by collecting data while changing the body states (the posture of the upper body, calibration, shoes, etc.), this information is self-organized in $\bm{p}$.
  $\bm{h}$ is a predictive model that represents the state transition of $\bm{s}$ by $\bm{u}$, and the dynamics of the model can be modified by changing $\bm{p}$.

  In this study, DPMPB has 10 layers, which consist of four FC layers (fully-connected layers), two LSTM layers (long short-term memory layers), and four FC layers.
  The number of units is set to \{$N_s+N_u+N_p$, 200, 100, 30, 30 (number of units in LSTM), 30 (number of units in LSTM), 30, 100, 200, $N_s$\} (where $N_{\{s, u, p\}}$ is the dimensionality of $\{\bm{s}, \bm{u}, \bm{p}\}$).
  The activation function is hyperbolic tangent and the update rule is Adam \cite{kingma2015adam}.
  We also set $\bm{p}$ to be two-dimensional and the execution period of \equref{eq:dpnpb} is 5 Hz.
  The dimension of $\bm{p}$ should be slightly smaller than the expected changes in the body state, because too small a dimensionality will not represent the change in dynamics properly, and too large a dimensionality will make self-organization of $\bm{p}$ difficult.
}%
{%
  本研究におけるDPMPBのネットワーク構造を以下に示す.
  \begin{align}
    \bm{s}_{t+1} = \bm{h}(\bm{s}_{t}, \bm{u}_{t}, \bm{p}) \label{eq:dpnpb}
  \end{align}
  ここで, $t$は現在のタイムステップ, $\bm{s}$はロボットの状態を表す変数, $\bm{u}$は制御入力, $\bm{p}$はParametric Bias, $\bm{h}$はDPMPBのネットワークを表す.
  本研究では筋骨格ヒューマノイドにおけるバランス制御を扱う.
  その中でも, 足首に関する関節や筋の状態を直接扱い, 上半身の姿勢等はParametric Biasによって暗黙的に扱う.
  そのため, $\bm{s}^{T} = \begin{pmatrix}\bm{z}_{t+1} & \bm{f}_{t+1} & \bm{l}_{t+1}\end{pmatrix}$, $\bm{u} = \theta^{ref}$とする.
  ここで, $\bm{z}$はZMPの値, $\bm{f}$は両脚の足首に関する筋張力, $\bm{l}$は両脚の足首に関する筋長, $\theta^{ref}$は足首の指令関節角度とする.
  なお, $\bm{z}$は2次元($x$方向は$z_x$, $y$方向は$z_y$とする)であり, $\bm{f}$や$\bm{l}$はロボットによって異なる.
  $\theta^{ref}$は両脚についてロール・ピッチの角度が考えられるが, 本研究では簡単のため両脚の角度を同一として, ピッチ軸のみの1次元とする.
  Parametric Bias $\bm{p}$は暗黙的なダイナミクスの違いを埋め込むことができる変数であり, 本研究では身体状態, 特に上半身の姿勢やキャリブレーション, 靴等を変化させながらデータを取得することで, これらの情報が$\bm{p}$の中に自己組織化される.
  $\bm{h}$は$\bm{u}$による$\bm{s}$の状態遷移を表現した予測モデルであり, $\bm{p}$を変化させることでそのモデルのダイナミクスを変化させることが可能である.

  本研究ではDPMPBは10層としており, 順に4つのFC層(fully-connected layer), 2層のLSTM層(long short-term memory), 4層のFC層からなる.
  ユニット数については, \{$N_s+N_u+N_p$, 200, 100, 30, 30 (LSTMのunit数), 30 (LSTMのunit数), 30, 100, 200, $N_s$\}とした(なお, $N_{\{s, u, p\}}$は$\{\bm{s}, \bm{u}, \bm{p}\}$の次元数とする).
  活性化関数はHyperbolic Tangent, 更新則はAdam \cite{kingma2015adam}とした.
  また, $\bm{p}$は2次元, \equref{eq:dpnpb}の実行周期は5 Hzとしている.
  $\bm{p}$の次元数は小さ過ぎるとdynamicsの変化を適切に表せなくなり, 大きすぎると自己組織化が難しくなるため, 想定される身体状態変化よりも少し小さく設定することが望ましい.
}%

\subsection{Data Collection} \label{subsec:data-collection}
\switchlanguage%
{%
  In order to learn balance control, some technique is needed in the method of data collection.
  If we simply move the ankles randomly, the robot will quickly fall down and it is difficult to collect useful data for balance control.
  In this study, $\theta^{ref}$ is varied by repeating the following process at each step,
  \begin{align}
    c &\gets c + 1\\
    d &\gets d + C_{diff}\\
    \theta^{ref} &\gets \theta^{ref} + |\sin(\pi\frac{c}{N_{cnt}})|\textrm{Random}(-d, d) \label{eq:collect}\\
    \theta^{ref} &\gets \max(\theta_{min}, \min(\theta^{ref}, \theta_{max}))
  \end{align}
  where $c$ is the time count (starting from $c=0$), $d$ is the maximum displacement of $\theta^{ref}$ (starting from $d=C^{init}_{diff}$), and $\textrm{Random}(a, b)$ is a random value in the range of $[a, b]$.
  Also, $\theta_{\{min, max\}}$ is \{minimum, maximum\} value of $\theta^{ref}$, and $\{N_{cnt}, C_{diff}, C^{init}_{diff}\}$ is a constant that determines the behavior of data collection.
  It collects data while gradually increasing the maximum value of the displacement of $\theta^{ref}$ with $d$.
  This is important because if the displacement is too large at the beginning, it will quickly fall down and we will not be able to collect data for a long time.
  Also, by periodically decreasing or increasing the change of $\theta^{ref}$ with $c$, we can collect various data.
  Since the best state for balance control is a stable stationary state, if we do not collect data for stationary states where the displacement of $\theta^{ref}$ is small, oscillatory motions will be generated during balance control.
  Finally, $\bm{\theta}^{ref}$ is clipped by the set minimum and maximum values.

  In the experiments, in addition to the data collection by \equref{eq:collect} (Proposed Collection), the following two types of data collection are compared,
  \begin{align}
    \theta^{ref} &\gets \theta^{ref} + \textrm{Random}(-d, d) \label{eq:collect-gradual}\\
    \theta^{ref} &\gets \theta^{ref} + \textrm{Random}(-1.0, 1.0) \label{eq:collect-random}
  \end{align}
  where we call \equref{eq:collect-gradual} Gradual Collection and \equref{eq:collect-random} Random Collection.
  Gradual Collection is a collection method excluding the periodic change of $\bm{\theta}^{ref}$ from Proposed Collection, and Random Collection is a collection method excluding the gradual increase of $\bm{\theta}^{ref}$ from Gradual Collection.

  In this study, we set $N_{cnt}=50$, $C_{diff}=0.002$ [rad], $C^{init}_{diff}=0.1$ [rad], $\theta_{min}=-1.0$ [rad], and $\theta_{max}=1.0$ [rad].
  Since the body is very difficult to modelize, some experimental tuning of these coefficients is necessary.
}%
{%
  本研究はバランス制御の学習であるため, そのデータ収集の方法には工夫が必要である.
  単にランダムに足首を動かすだけではすぐに倒れてしまうし, バランス制御に有用なデータを集めにくい.
  本研究では以下の操作を毎ステップ繰り返すことにより$\theta^{ref}$を変化させていく.
  \begin{align}
    c &\gets c + 1\\
    d &\gets d + C_{diff}\\
    \theta^{ref} &\gets \theta^{ref} + |\sin(\pi\frac{c}{N_{cnt}})|\textrm{Random}(-d, d) \label{eq:collect}\\
    \theta^{ref} &\gets \max(\theta_{min}, \min(\theta^{ref}, \theta_{max}))
  \end{align}
  ここで, $c$はタイムカウント($c=0$から始める), $d$は$\theta^{ref}$の変位の最大値($d=C^{init}_{diff}$から始める), $\textrm{Random}(a, b)$は$[a, b]$の範囲のランダムな値を表す.
  また, $\theta_{\{min, max\}}$は$\theta^{ref}$の最小値最大値, $N_{cnt}$, $C_{diff}$, $C^{init}_{diff}$は挙動を決める定数である.
  これは, $\theta^{ref}$の変位の最大値を$d$を上げることで徐々に増やしながらデータを収集する.
  初めから変位が大きいとすぐに倒れてしまい長くデータが取れないので, これは重要である.
  また, $\theta^{ref}$の変化を$c$によって周期的に小さくしたり大きくしたりすることで, 多様なデータを収集する.
  バランス制御にとって最も良い状態は安定して止まる状態なため, $\theta^{ref}$の変位が小さい止まった状態のデータも収集しないと振動的な動作が生成されてしまうことになる.
  最後に, $\bm{\theta}^{ref}$を設定した最小値と最大値によりclipする.

  実験においては, \equref{eq:collect}によるデータ収集(Proposed Collection)に加え, 以下の2種類のデータ収集を比較する.
  \begin{align}
    \theta^{ref} &\gets \theta^{ref} + \textrm{Random}(-d, d) \label{eq:collect-gradual}\\
    \theta^{ref} &\gets \theta^{ref} + \textrm{Random}(-1.0, 1.0) \label{eq:collect-random}
  \end{align}
  ここで, \equref{eq:collect-gradual}をGradual Collection, \equref{eq:collec-random}をRandom Collectionと呼ぶ.
  Gradual CollectionはProposed Collectionから周期的な$\bm{\theta}^{ref}$の変化を除いたもの, Random CollectionはGradual Collectionから$\bm{\theta}^{ref}$の変位の段階的な上昇を除いたものである.

  本研究では$N_{cnt}=50$, $C_{diff}=0.002$ [rad], $C^{init}_{diff}=0.1$ [rad], $\theta_{min}=-1.0$ [rad], $\theta_{max}=1.0$ [rad]とする.
  身体のモデル化が非常に難しい系であるため, これらの係数についてはある程度の実験的なtuningが必要である.
}%

\subsection{Training of DPMPB}
\switchlanguage%
{%
  Using the obtained data $D$, DPMPB is trained.
  In this process, we can implicitly embed the information of body state into parametric bias by collecting data while changing the body state.
  In order to allow each time-series data transition with different dynamics to be represented by a single model, the differences in the dynamics is self-organized in a low-dimensional space of $\bm{p}$.
  It can be regarded as a weakly supervised learning, in which only weak labels are given, i.e., whether or not the body state is the same for each data.

  The data collected in the same body state $k$ is represented as $D_{k}=\{(\bm{s}_{1}, \bm{u}_{1}), (\bm{s}_{2}, \bm{u}_{2}), \cdots, (\bm{s}_{T_{k}}, \bm{u}_{T_{k}})\}$ ($1 \leq k \leq K$, where $K$ is the total number of body states and $T_{k}$ is the number of motion steps for the body state $k$), and the data used for training $D_{train}=\{(D_{1}, \bm{p}_{1}), (D_{2}, \bm{p}_{2}), \cdots, (D_{K}, \bm{p}_{K})\}$ is constructed.
  Here, $\bm{p}_{k}$ is the parametric bias that represents the dynamics in the body state $k$, which is a common variable for that state and a different variable for another state.
  We use $D_{train}$ to train the DPMPB.
  In an ordinary learning process, only the network weight $W$ is updated, but here, $W$ and $p_{k}$ for each state are updated simultaneously.
  In this way, $p_{k}$ embeds the difference of dynamics in each body state.
  In the learning process, the mean squared error is used as the loss function, and $\bm{p}_{k}$ is optimized with all initial values set to $\bm{0}$.
}%
{%
  得られたデータ$D$を使いDPMPBを学習させる.
  この際, 身体状態を変化させながらデータを収集することで, これらのデータを暗黙的にParametric Biasに埋め込むことができる.
  異なるダイナミクスを持つそれぞれの時系列データ遷移を一つのモデルで表現できるように, そのダイナミクスの違いを低次元の$\bm{p}$の空間に形成する.
  それぞれのデータについて身体状態が同一かどうかという弱い教師のみを与えた弱教師あり学習と捉えることも可能である.

  ある同一の身体状態$k$において収集されたデータを$D_{k}=\{(\bm{s}_{1}, \bm{u}_{1}), (\bm{s}_{2}, \bm{u}_{2}), \cdots, (\bm{s}_{T_{k}}, \bm{u}_{T_{k}})\}$ ($1 \leq k \leq K$, $K$は全試行回数, $T_{k}$はその身体状態$k$における試行の動作ステップ数)として, 学習に用いるデータ$D_{train}=\{(D_{1}, \bm{p}_{1}), (D_{2}, \bm{p}_{2}), \cdots, (D_{K}, \bm{p}_{K})\}$を得る.
  ここで, $\bm{p}_{k}$はその身体状態$k$におけるダイナミクスを表現するParametric Biasであり, その状態については共通の変数, 別の状態については別の変数となる.
  この$D_{train}$を使ってDPMPBを学習させる.
  通常の学習ではネットワークの重み$W$のみが更新されるが, ここでは$W$と各状態に関する$p_{k}$が同時に更新される.
  これにより, $p_{k}$にそれぞれの身体状態におけるダイナミクスの違いが埋め込まれることになる.
  学習の際は損失関数として平均二乗誤差を使い, $\bm{p}_{k}$は全て0を初期値として最適化される.
}%

\subsection{Online Update of Parametric Bias}
\switchlanguage%
{%
  Using the data $D$ obtained in the current body state, we update parametric bias online.
  If the network weight $W$ is updated, DPMPB may overfit to the data, but if only the low-dimensional parametric bias $\bm{p}$ is updated, no overfitting occurs and life-long update is possible.
  Note that it has been experimented in \cite{kawaharazuka2021tooluse} that fine tuning of only $W$ without using $\bm{p}$ cannot deal with various body states, though this is a study on a static motion model.
  This online learning allows us to obtain a model that is always adapted to the current body state.

  Let the number of data obtained be $N^{online}_{data}$, and start online learning when the number of data exceeds $N^{online}_{thre}$.
  For each new data, we fix $W$ and update only $\bm{p}$ by setting the number of batches as $N^{online}_{batch}$, the number of epochs as $N^{online}_{epoch}$, and the update rule as MomentumSGD.
  Data exceeding $N^{online}_{max}$ are deleted from the oldest ones.

  In this study, we set $N^{online}_{\{thre, max\}}=50$, $N^{online}_{batch}=N^{online}_{max}$, and $N^{online}_{epoch}=1$.
}%
{%
  現在の身体状態において得られたデータ$D$を使い, オンラインでParametric Biasを更新する.
  ネットワークの重み$W$を更新してしまうとDPMPBがそのデータに過学習してしまう可能性があるが, 低次元のParametric Bias $\bm{p}$のみを更新するのであれば過学習は起こらず, 常に学習し続けることが可能になる.
  なお, 静的な動作モデルに関する研究であるが, $\bm{p}$を用いずに$W$のみをFine Tuningする方法では異なる身体状態に対応できないことは\cite{kawaharazuka2021tooluse}において実験されている.
  このオンライン学習により, 常に現在の身体状態に適応したモデルを得ることができる.

  得られたデータ数を$N^{online}_{data}$として, データ数が$N^{online}_{thre}$を超えたところからオンライン学習を始める.
  新しいデータが入るたびにバッチ数を$N^{online}_{batch}$, エポック数を$N^{online}_{epoch}$, 更新則をMomentumSGDとして学習を行う.
  $N^{online}_{max}$を超えたデータは古いものから削除していく.

  本研究では, $N^{online}_{\{thre, max\}}=50$, $N^{online}_{batch}=N^{online}_{max}$, $N^{online}_{epoch}=1$とした.
}%

\subsection{Balance Control using DPMPB}
\switchlanguage%
{%
  We describe a control method using DPMPB.
  Here, we consider optimizing $\bm{u}$ from the loss function for $\bm{s}$ and $\bm{u}$.
  First, we give the initial value $\bm{u}^{init}_{seq}$ for the time-series control input $\bm{u}_{seq}=\bm{u}_{[t:t+N_{step}-1]}$ ($N_{step}$ represents the number of DPMPB expansions, or control horizon).
  Let $\bm{u}^{opt}_{seq}$ be $\bm{u}$ to be optimized, and repeat the following calculation at the time step $t$ to obtain the optimal $\bm{u}^{opt}_{t}$,
  \begin{align}
    \bm{s}^{pred}_{seq} &= \bm{h}_{expand}(\bm{s}_{t}, \bm{u}^{opt}_{seq})\\
    L &= \bm{h}_{loss}(\bm{s}^{pred}_{seq}, \bm{u}^{opt}_{seq}) \label{eq:control-loss}\\
    \bm{u}^{opt}_{seq} &\gets \bm{u}^{opt}_{seq} - \gamma\partial{L}/\partial{\bm{u}^{opt}_{seq}} \label{eq:control-opt}
  \end{align}
  where $\bm{s}^{pred}_{seq}$ is the predicted $\bm{s}_{[t+1:t+N_{step}]}$, $\bm{h}_{expand}$ is the function of $\bm{h}$ expanded $N_{step}$ times, $\bm{h}_{loss}$ is the loss function, and $\gamma$ is the learning rate.
  Thus, the future $\bm{s}$ is predicted from the current sensor state $\bm{s}_{t}$ by $\bm{u}^{opt}_{seq}$, and $\bm{u}^{opt}_{seq}$ is optimized by using the backpropagation and gradient descent methods to minimize the loss function.

  In this process, we set $\bm{u}^{init}_{seq}$ as $\bm{u}^{prev}_{\{t+1, \cdots, t+N_{step}-1, t+N_{step}-1\}}$ by using $\bm{u}^{prev}_{[t:t+N_{step}-1]}$, which is $\bm{u}_{seq}$ optimized in the previous step, shifting the time by one, and replicating the last term.
  By using the previous optimization result, faster convergence can be obtained.
  For $\gamma$, we prepare $N^{control}_{batch}$ number of $\gamma$, which are exponentially divided $[0, \gamma_{max}]$, and after running \equref{eq:control-opt} on each $\gamma$, we calculate \equref{eq:control-loss} and select the $\bm{u}^{opt}_{seq}$ with the lowest loss, repeating the process $N^{control}_{epoch}$ times.
  Faster convergence can be obtained by trying various $\gamma$ and always choosing the best learning rate.

  Here, we consider the loss function.
  In this study, we set $\bm{h}_{loss}$ as follows,
  \begin{align}
    \bm{h}_{loss}(\bm{s}^{pred}_{seq}, \bm{u}^{opt}_{seq}) &= ||\bm{z}^{pred}_{seq}-\bm{z}^{ref}_{seq}||_{2}\nonumber\\
    &+ C_{f}||\bm{f}^{pred}_{[3:N_{step}]}-\bm{f}^{pred}_{[2:N_{step}-1]}||_{2}\nonumber\\
    &+ C_{l}||\bm{l}^{pred}_{[3:N_{step}]}-\bm{l}^{pred}_{[2:N_{step}-1]}||_{2}\nonumber\\
    &+ C_{u}||\bm{u}^{opt}_{seq}||_{2} \label{eq:loss}
  \end{align}
  where $\{\bm{z}, \bm{f}, \bm{l}\}^{pred}_{seq}$ is the value of $\{\bm{z}, \bm{f}, \bm{l}\}$ in $\bm{s}^{pred}_{seq}$, $\bm{z}^{ref}_{seq}$ is the value obtained by arranging $N_{step}$ target values of $\bm{z}$, and $C_{\{f, l, u\}}$ is the constant weight for each loss.
  Thus, the loss is a summary of the realization of the target value for $\bm{z}$, the minimization of the change in $\bm{f}$, the minimization of the change in $\bm{l}$, and the minimization of $\bm{u}$.
  Note that $C_{\{f, l, u\}}$ is varied for each experiment.

  In this study, we set $N_{step}=6$, $N^{control}_{batch}=10$, $N^{control}_{epoch}=3$, and $\gamma_{max}=0.1$.
}%
{%
  DPMPBを使った制御手法について述べる.
  ここでは, $\bm{s}$と$\bm{u}$に関する損失関数から, $\bm{u}$を最適化していくことを考える.
  まず, 時系列制御入力$\bm{u}_{seq}=\bm{u}_{[t:t+N_{step}-1]}$の初期値$\bm{u}^{init}_{seq}$を与える($N_{step}$は制御におけるDPMPBの展開数, 制御ホライズンを表す).
  最適化する$\bm{u}$を$\bm{u}^{opt}_{seq}$とおき, 時刻$t$において以下の計算を繰り返すことで最適な$\bm{u}^{opt}_{t}$を得る.
  \begin{align}
    \bm{s}^{pred}_{seq} &= \bm{h}_{expand}(\bm{s}_{t}, \bm{u}^{opt}_{seq})\\
    L &= \bm{h}_{loss}(\bm{s}^{pred}_{seq}, \bm{u}^{opt}_{seq}) \label{eq:control-loss}\\
    \bm{u}^{opt}_{seq} &\gets \bm{u}^{opt}_{seq} - \gamma\partial{L}/\partial{\bm{u}^{opt}_{seq}} \label{eq:control-opt}
  \end{align}
  ここで, $\bm{s}^{pred}_{seq}$は予測された$\bm{s}_{[t+1:t+N_{step}]}$, $\bm{h}_{expand}$は$\bm{h}$を$N_{step}$回展開した関数, $\bm{h}_{loss}$は損失関数, $\gamma$は学習率を表す.
  つまり, 現在状態$\bm{s}_{t}$から時系列制御入力$\bm{u}^{opt}_{seq}$により将来の$\bm{s}$を予測し, これに対して設定した損失関数を最小化するように, $\bm{u}^{opt}_{seq}$を誤差逆伝播法と勾配法により最適化する.

  このとき$\bm{u}^{init}_{seq}$は, 前ステップで最適化された$\bm{u}_{seq}$である$\bm{u}^{prev}_{[t:t+N_{step}-1]}$を使い, 時間を一つシフトし最後の項を複製した$\bm{u}^{prev}_{\{t+1, \cdots, t+N_{step}-1, t+N_{step}-1\}}$とする.
  前回の最適化結果を利用することでより速い収束が得られる.
  また, $\gamma$については$[0, \gamma_{max}]$を指数関数的に分割した$N^{control}_{batch}$個の$\gamma$を用意し, それぞれの$\gamma$で\equref{eq:control-opt}を実行した後, \equref{eq:control-loss}で損失を計算し最も損失の小さい$\bm{u}^{opt}_{seq}$を選択することを$N^{control}_{epoch}$回繰り返す.
  様々な$\gamma$を試して最良の学習率を常に選ぶことで, より速い収束が得られる.

  ここで, 損失関数$\bm{h}_{loss}$について考える.
  本研究では$\bm{h}_{loss}$を以下のように設定する.
  \begin{align}
    \bm{h}_{loss}(\bm{s}^{pred}_{seq}, \bm{u}^{opt}_{seq}) &= ||\bm{z}^{pred}_{seq}-\bm{z}^{ref}_{seq}||_{2}\nonumber\\
    &+ C_{f}||\bm{f}^{pred}_{[3:N_{step}]}-\bm{f}^{pred}_{[2:N_{step}-1]}||_{2}\nonumber\\
    &+ C_{l}||\bm{l}^{pred}_{[3:N_{step}]}-\bm{l}^{pred}_{[2:N_{step}-1]}||_{2}\nonumber\\
    &+ C_{u}||\bm{u}^{opt}_{seq}||_{2} \label{eq:loss}
  \end{align}
  ここで, $\{\bm{z}, \bm{f}, \bm{l}\}^{pred}_{seq}$は$\bm{s}^{pred}_{seq}$における$\{\bm{z}, \bm{f}, \bm{l}\}$の値, $\bm{z}^{ref}_{seq}$は$\bm{z}$の指令値を$N_{step}$個並べた値, $C_{\{f, l, u\}}$はそれぞれの損失の重みの定数を表す.
  つまり, $\bm{z}$に関する指令値の実現, $\bm{f}$の変化の最小化, $\bm{l}$の変化の最小化, $\bm{u}$の最小化をまとめた損失となっている.
  なお, $C_{\{f, l, u\}}$は実験ごとに変化させる.

  本研究では, $N_{step}=6$, $N^{control}_{batch}=10$, $N^{control}_{epoch}=3$, $\gamma_{max}=0.1$とする.
}%

\begin{figure}[t]
  \centering
  \includegraphics[width=0.5\columnwidth]{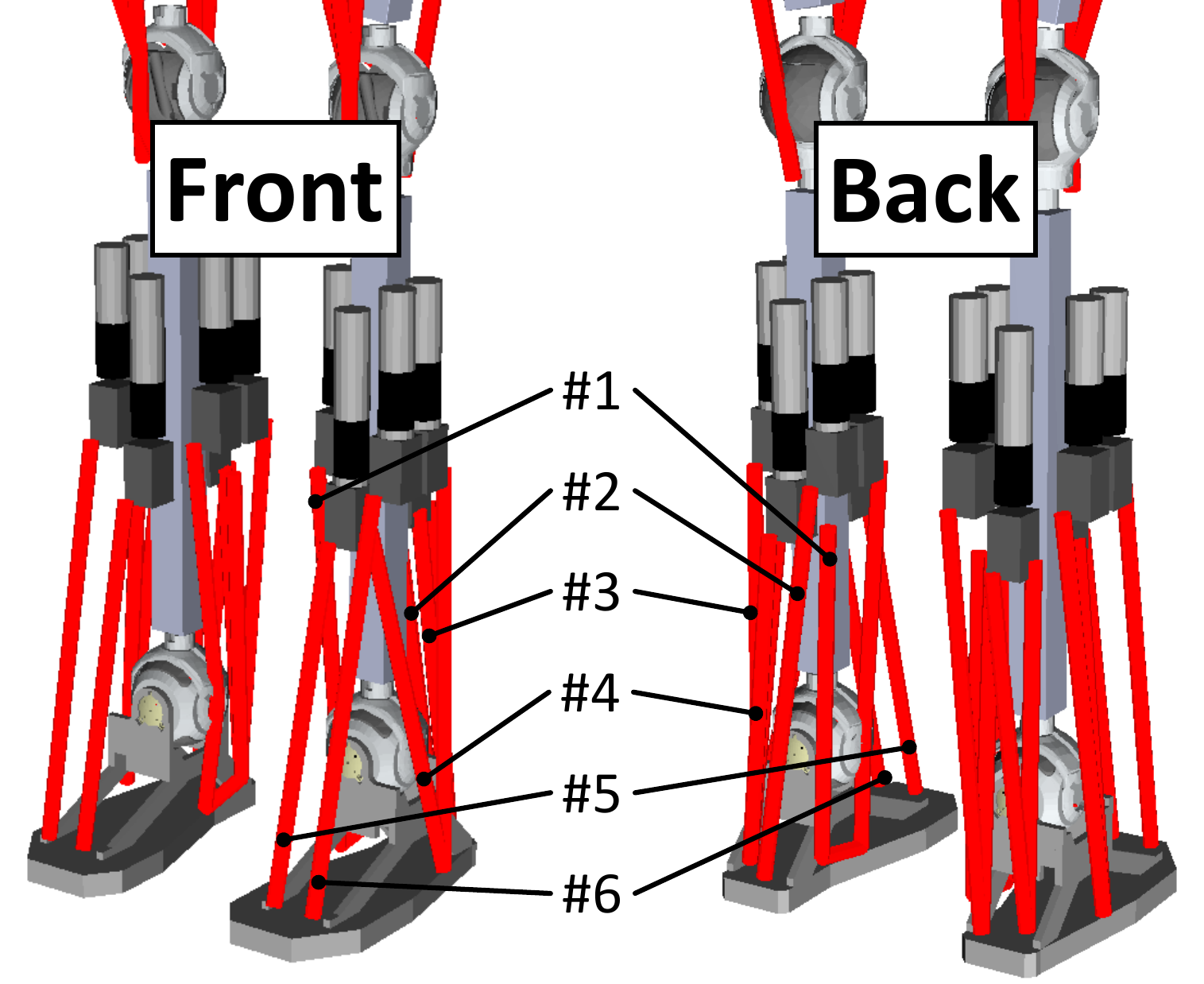}
  \vspace{-1.0ex}
  \caption{The muscle arrangement of the musculoskeletal humanoid Musashi.}
  \label{figure:exp-setup}
  \vspace{-1.0ex}
\end{figure}

\section{Experiments} \label{sec:experiment}

\subsection{Experimental Setup}
\switchlanguage%
{%
  In this study, we conduct experiments using the musculoskeletal humanoid Musashi (\figref{figure:concept}) \cite{kawaharazuka2019musashi}.
  Musashi has redundant 74 muscles including 4 polyarticular muscles in its body and 34 over-actuated joints excluding fingers.
  In this study, Musashi basically moves only the pitch joint of the ankle and controls balance in an upright posture except for in some upper body postures.
  As shown in \figref{figure:exp-setup}, there are six muscles for each ankle joint, and the dimensionality of $\{\bm{f}, \bm{l}\}$ for both legs is 12.
  ZMP is calculated from 12 loadcells distributed in the foot.
  \cite{kawaharazuka2019longtime} is used to convert the target joint angle to target muscle length, assuming the target muscle tension to be constant at 100 [N].
  Note that the learning of \cite{kawaharazuka2019longtime} does not perfectly realize the target joint angle, and there are some errors due to muscle friction and other factors.
  For simulation, we use Mujoco \cite{todorov2012mujoco}.
}%
{%
  本研究では筋骨格ヒューマノイドMusashi \cite{kawaharazuka2019musashi}を用いて実験を行う.
  Musashiは全身に74本の筋(うち4本が多関節筋)を有しており, 指を除いた関節数は34, 劣駆動関節はない.
  本研究では基本的に足首のピッチ関節のみを動かし, 一部上半身を動かす以外については, 直立した姿勢でバランス制御を行う.
  \figref{figure:exp-setup}の筋配置に示すように, 足首の筋は6本であり, 両脚合わせて$\bm{f}$と$\bm{l}$は12次元である.
  ZMPについては, 足平に分布する12個のロードセルから計算している.
  また, 筋骨格ヒューマノイドにおいては指令関節角度を筋長に変換する必要があるが, 本研究では\cite{kawaharazuka2019longtime}を使い, 指令筋張力を100 [N]で一定として変換している.
  なお, \cite{kawaharazuka2019longtime}の学習は完璧に指令関節角度を実現できるわけではなく, 筋の摩擦等によって誤差がのる.
  シミュレーションについてはMujoco \cite{todorov2012mujoco}を使う.
}%

\begin{figure}[t]
  \centering
  \includegraphics[width=0.98\columnwidth]{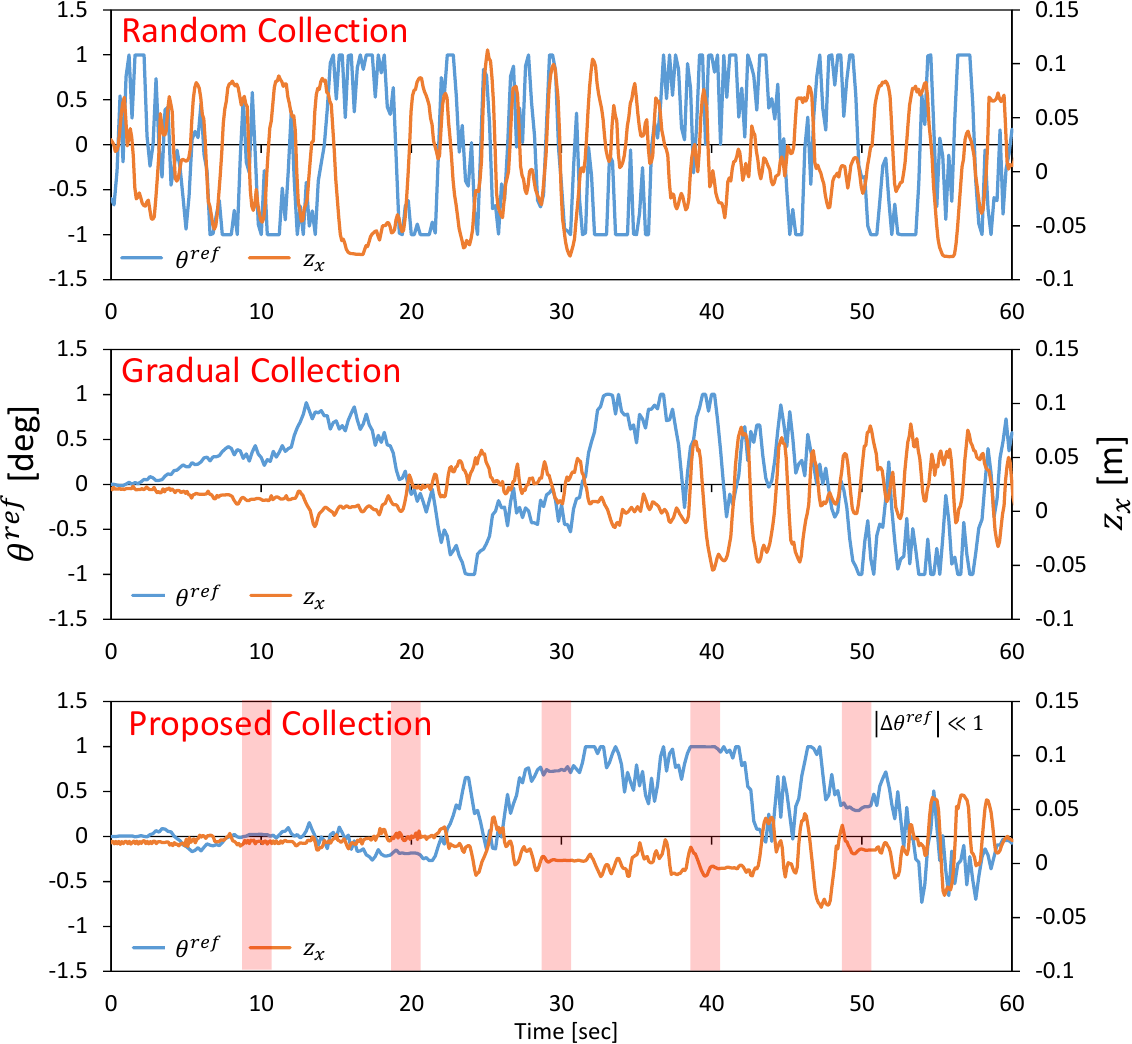}
  \vspace{-1.0ex}
  \caption{Simulation experiment: the transition of $\theta^{ref}$ and $z_x$ when conducting Random, Gradual, and Proposed Collections.}
  \label{figure:sim-collect}
  \vspace{-1.0ex}
\end{figure}

\begin{figure*}[t]
  \centering
  \includegraphics[width=1.98\columnwidth]{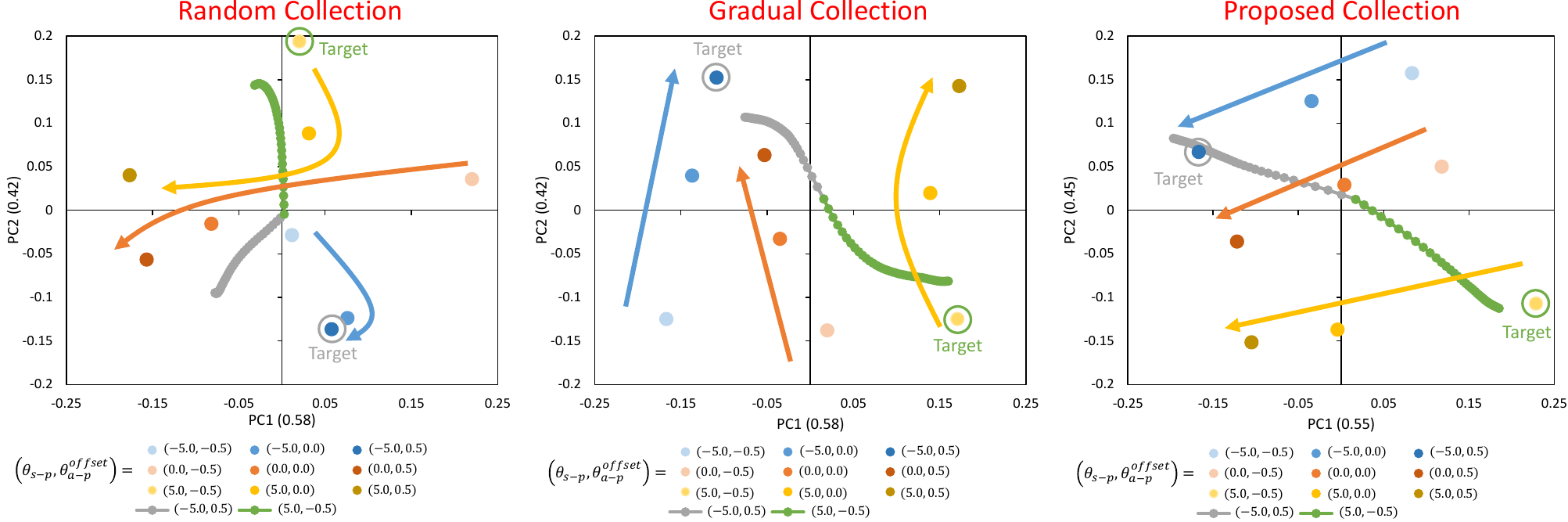}
  \vspace{-1.0ex}
  \caption{Simulation experiment: the arrangement of parametric bias when training DPMPB using the data collected with Random, Gradual, and Proposed Collection, and the trajectories of parametric bias when running online learning by setting $(\theta_{s-p}, \theta^{offset}_{a-p}) = \{(-5.0, 0.5), (5.0, -0.5)\}$.}
  \label{figure:sim-pb}
\end{figure*}

\begin{figure*}[t]
  \centering
  \includegraphics[width=1.98\columnwidth]{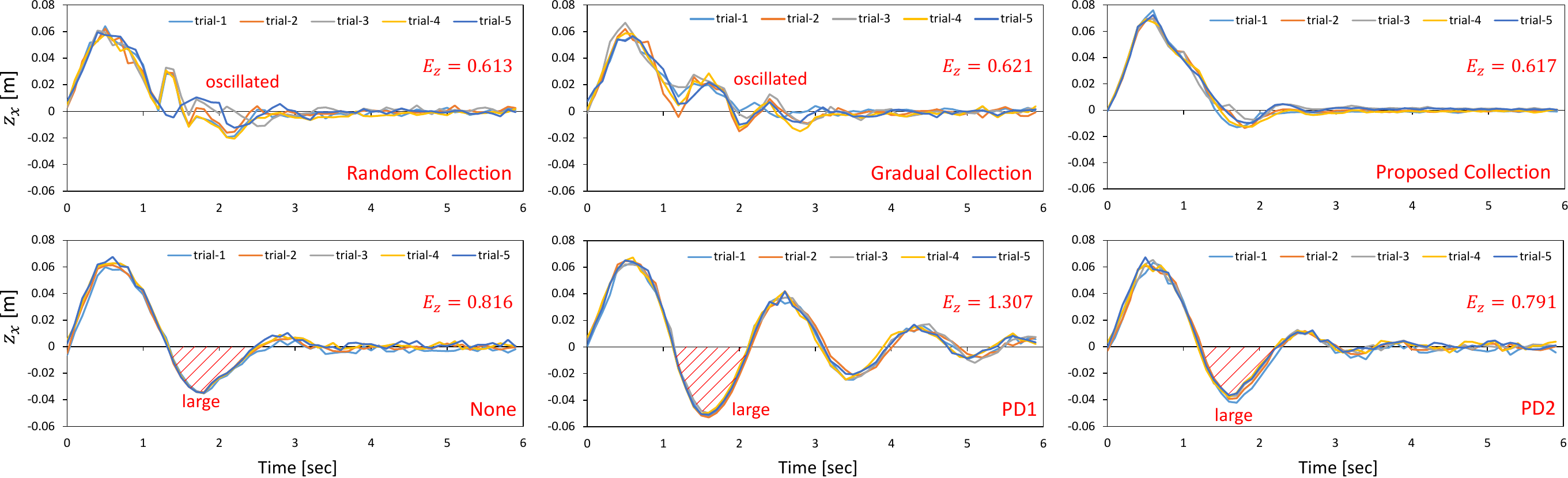}
  \vspace{-1.0ex}
  \caption{Simulation experiment: the transitions of $z_x$ after applying 30 N force to the chest link for 0.2 seconds, when the balance control using DPMPB trained with the data collected by Random, Gradual, or Proposed Collection is performed, when no balance control is performed (None), or when simple PD controls with different gains are performed (PD1, PD2).}
  \label{figure:sim-control1}
\end{figure*}

\begin{figure}[t]
  \centering
  \includegraphics[width=0.98\columnwidth]{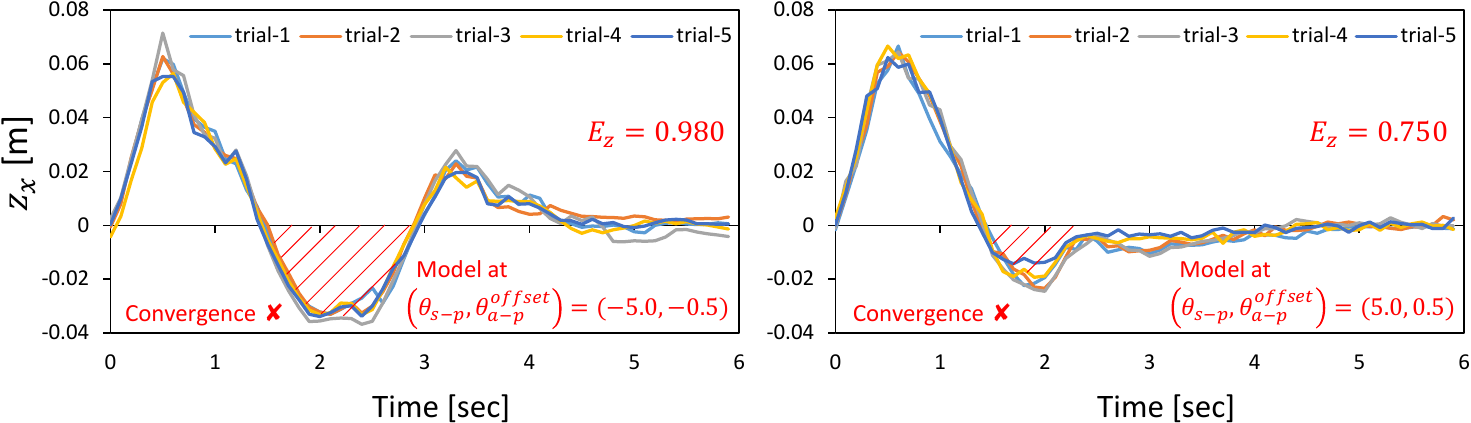}
  \vspace{-1.0ex}
  \caption{Simulation experiment: the transitions of $z_x$ after external force of 30 N to the chest link for 0.2 seconds when running the balance control using DPMPB trained with the data collected by Proposed Collection and parametric bias trained at $(\theta_{s-p}, \theta^{offset}_{a-p})=\{(-5.0, -0.5), (5.0, 0.5)\}$.}
  \label{figure:sim-control2}
  \vspace{-1.0ex}
\end{figure}

\begin{figure*}[t]
  \centering
  \includegraphics[width=1.98\columnwidth]{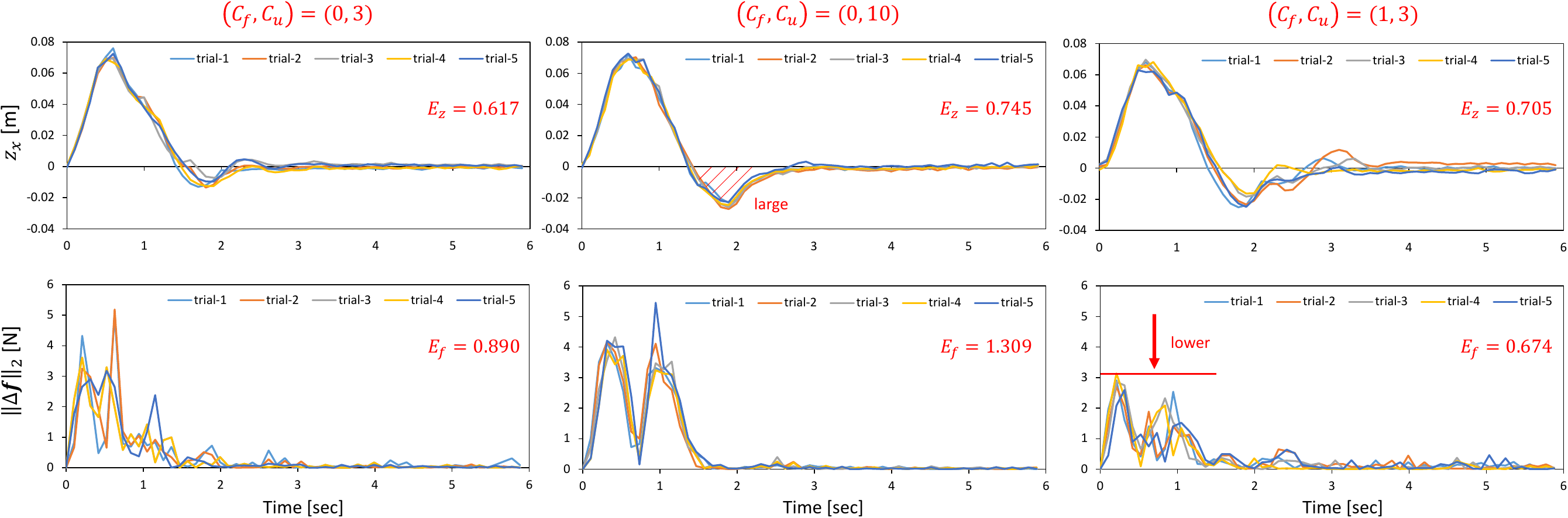}
  \vspace{-1.0ex}
  \caption{Simulation experiment: the transitions of $z_x$ and $||\Delta{f}||_{2}$ when running the proposed balance control with $(C_f, C_u) = \{(0, 3), (0, 10), (1, 3)\}$ after external force of 30 N is applied to the chest link for 0.2 seconds.}
  \label{figure:sim-control3}
  \vspace{-1.0ex}
\end{figure*}

\subsection{Simulation Experiment}
\switchlanguage%
{%
  \subsubsection{Training of DPMPB}
  In this experiment, we handle the pitch angle of the spine joint $\theta_{s-p}$, and the offset of the pitch angle of the ankle joint $\theta^{offset}_{a-p}$ representing irreproducible calibration.
  First, we collect data while changing the body state to nine combinations of $\theta_{s-p}=\{-5.0, 0.0, 5.0\}$ [deg] and $\theta^{offset}_{a-p}=\{-5.0, 0.0, 5.0\}$ [deg].
  For each body state, we obtain data for 300 time steps.
  Here, the transitions of $z_x$ and $\theta^{ref}$ are shown in \figref{figure:sim-collect} when using Proposed, Gradual, or Random Collection in \secref{subsec:data-collection}.
  In Random Collection, $z_x$ and $\theta^{ref}$ continue to change significantly.
  In Gradual Collection, the range of change in $z_x$ and $\theta^{ref}$ gradually increases.
  On the other hand, in Proposed Collection, in addition to the characteristics of Gradual Collection, $\theta^{ref}$ alternates between violent and slow motions, and a variety of data is collected.
  We train DPMPB using the data of 2700 time steps.
  For each data collection method, the obtained parametric bias is converted by Principle Component Analysis (PCA) and plotted on a two-dimensional plane as shown in \figref{figure:sim-pb}.
  In Proposed Collection, we can see that the space of PB is self-organized according to the size of the body state parameters, though the parameters related to the body state are not directly given as data.
  In other words, even in the case where the parameters of the body state are not directly available, such as in the recalibration of the actual robot, it is possible to structure the information in the space of PB.
  On the other hand, Gradual Collection shows a more distorted space of PBs than Proposed Collection.
  As for Random Collection, the space of PB is even more distorted than that of Gradual Collection.

  \subsubsection{Online Update of Parametric Bias}
  Starting from the state of $\bm{p}=\bm{0}$, we examine how $\bm{p}$ transitions when the online update of PB is performed at the same time as when the body is moved the same way as in the data collection.
  The trajectory of $\bm{p}$ when $(\theta_{s-p}, \theta^{offset}_{a-p})=\{(-5.0, 0.5), (5.0, -0.5)\}$ is shown in \figref{figure:sim-pb}.
  Note that the trajectories for 45 online learning steps are shown.
  It can be seen that the current $\bm{p}$ is gradually approaching the $\bm{p}$ previously trained in the same body state as the current state.
  In other words, it is possible to correctly recognize the body state by searching the space of $\bm{p}$.
  In addition, the accuracy of the recognition increases in the order of Random $<$ Gradual $<$ Proposed Collection.

  \subsubsection{Balance Control Using DPMPB}
  In this experiment, we set $(\theta_{s-p}, \theta^{offset}_{a-p})=(0.0, 0.0)$, and the transition of $z_x$ after applying an external force of 30N to the waist link for 0.2 seconds is examined five times for 6 seconds each.
  For $z_x$, offsets are removed to align the origins of the plots, and the average of the sum of $|z_x|$ for 6 seconds (30 steps) is shown as $E_z$.
  Unless otherwise stated, the constant weight for the loss function is set to $(C_{f}, C_{l}, C_{u})=(0, 30, 3)$, and PB is the value obtained when $(\theta_{s-p}, \theta^{offset}_{a-p})=(0.0, 0.0)$.

  First, we show the results for the cases of balance control using the models obtained for Random, Gradual, and Proposed Collections, no control (None), and PD control (PD), in \figref{figure:sim-control1}.
  As examples of PD controls, we show the cases of $(K_P, K_D)=(0.1, 0.1)$ (PD1) and $(K_P, K_D)=(0.03, 0.1)$ (PD2), though any PD setting would have worked to prevent the convergence of $z_x$ ($K_{\{P, D\}}$ is the gain for PD control).
  Note that PD2 is the best controller tuned manually but it cannot be denied that tuning methods such as \cite{fiducioso2019pid} may somewhat improve the results.
  It can be seen that the models using Random Collection and Gradual Collection do not change $E_f$ significantly compared to the model using Proposed Collection, but $z_x$ becomes oscillatory.
  In the case of None, $E_z$ is larger than when using the control of this study, and $x_z$ swings once in the positive direction and then again in the negative direction.
  On the other hand, in the case of our control, $E_z$ does not swing much in the negative direction and converges faster.
  In the case of PD, even if the gain is changed, the convergence becomes worse than None in most cases.

  Next, the results for the case where the model obtained in Proposed Collection is used and PB is the value obtained when $(\theta_{s-p}, \theta^{offset}_{a-p})=\{(-5.0, -0.5), (5.0, 0.5)\}$ are shown in \figref{figure:sim-control2}.
  For both cases, we can see that the error is much larger than that when using the correct PB obtained at $(\theta_{s-p}, \theta^{offset}_{a-p})=(0.0, 0.0)$.

  Finally, for the case of the model obtained in Proposed Collection, $C_{l}=30$ is fixed and the parameters of the balance control are varied as $(C_{f}, C_{u})=\{(0, 3), (1, 3), (0, 10)\}$.
  The results are shown in \figref{figure:sim-control3}.
  The transition of the norm $||\Delta\bm{f}||_{2}$ of the time variation of the muscle tension from the previous step is also shown here, and the root mean square of the values is denoted by $E_f$.
  The upper left figure of \figref{figure:sim-control3} is the same graph as Proposed Collection of \figref{figure:sim-control1}.
  It can be seen that changing $C_{u}$ from 3 to 10 suppresses the movement of $\bm{u}=\theta^{ref}$, so that the movement of $z_{x}$ approaches None in \figref{figure:sim-control1}.
  It can also be seen that when $C_{f}$ is changed from 0 to 1, the peak of $||\Delta\bm{f}||_{2}$ subsides and $E_{f}$ becomes 0.674, which is smaller than in the case of $C_{f}=0$.
}%
{%
  シミュレーション実験を行う.
  ここでは身体状態変化として, 腰関節のピッチ軸角度$\theta_{s-p}$とキャリブレーションのズレを表現した足首ピッチ関節角度のオフセット$\theta^{offset}_{a-p}$を扱う.
  まずデータ収集を行うが, この際身体状態を$\theta_{s-p}=\{-5.0, 0.0, 5.0\}$ [deg], $\theta^{offset}_{a-p}=\{-5.0, 0.0, 5.0\}$ [deg]によって9種類に変化させながらデータを取る.
  それぞれの身体状態について, 300ステップのデータを得る.
  この際, \secref{subsec:data-collection}におけるProposed, Gradual, Random Collectionのそれぞれを用いた場合における$z_x$と$\theta^{ref}$の遷移を\figref{figure:sim-collect}に示す.
  Random Collectionでは常に$z_x$と$\theta^{ref}$が大きく変化する.
  Gradual Collectionでは$z_x$と$\theta^{ref}$の変化幅は徐々に大きくなっていく.
  一方Proposed CollectionではGradual Collectionの特徴に加えて, $\theta^{ref}$が激しく動くところとゆったり動くところが交互に訪れており, 多様なデータが収集されている.
  この2700ステップのデータを使ってDPMPBを学習させる.
  それぞれのデータ取得方法について, 得られたParametric BiasにPrinciple Component Analysis (PCA)をかけ, 2次元平面上にプロットしたものを\figref{figure:sim-pb}に示す.
  Proposed Collectionを見ると, 直接身体状態に関するパラメータはデータとして与えていないが, 綺麗に身体状態パラメータの大小に従って値が自己組織化していることがわかる.
  つまり, 実機の再キャリブレーションのような直接身体状態のパラメータが得られないような場合にも, これらをParametric Biasの空間に構造化することが可能である.
  一方で, Gradual CollectionはProposed Collectionに比べてPBの空間が歪んでいることがわかる.
  また, Random CollectionについてはGradual Collectionよりも更にPBの空間が歪んでいる.

  次に, Parametric Biasのオンライン更新について実験を行う.
  $\bm{p}=\bm{0}$の状態から始め, データ収集時と同様に体を動かすと同時にPBのオンライン更新を実行したときに, どのように$\bm{p}$が遷移するかを調べる.
  $(\theta_{s-p}, \theta^{offset}_{a-p})=\{(-5.0, 0.5), (5.0, -0.5)\}$としたときの$\bm{p}$の軌跡を\figref{figure:sim-pb}に示す.
  なお, オンライン学習45ステップ分の軌跡を表示している.
  徐々に現在の$\bm{p}$が, 現在と同じ身体状態において学習された$\bm{p}$へと近づいていくことがわかる.
  つまり, $\bm{p}$の空間を探索して身体状態を正しく認識するような適応が可能である.
  また, その精度はRandom $<$ Gradual $<$ Proposed Collectionの順で高くなる.

  最後に, 制御について実験を行う.
  本実験では全て$(\theta_{s-p}, \theta^{offset}_{a-p})=(0.0, 0.0)$とし, 腰リンクに対して30Nの外力を0.2秒間加えた後の$z_x$の遷移を6秒間, 5回ずつ検証する.
  $z_x$についてはグラフを揃えるためにオフセットを取り除き, 6秒間(30ステップ分)の$|z_x|$の合計値の平均を$E_z$として表示している.
  また記載がない限り, 制御については$(C_{f}, C_{l}, C_{u})=(0, 30, 3)$と設定し, PBについては$(\theta_{s-p}, \theta^{offset}_{a-p})=(0.0, 0.0)$の際に得られた値とする.

  まずは, Random, Gradual, Proposed Collectionにおいて得られたそれぞれのモデルを使ってバランス制御をした場合, 何も制御しなかった場合(None), PD制御をした場合(PD)の結果を\figref{figure:sim-control1}に示す.
  PDについてはどのような設定をしても$z_x$の収束を妨げる方向に働いてしまったが, 例として$(K_P, K_D)=(0.1, 0.1)$ (PD1), $(K_P, K_D)=(0.03, 0.1)$ (PD2)の場合を表示している(PD2は手動でtuningした最良のコントローラである. なお, \cite{fiducioso2019pid}等のtuning手法によって多少良くなる可能性は否定できない).
  Random CollectionとGradual Collectionのデータを使ったモデルはProposed Collectionのデータを使った場合に比べて, $E_f$は大きく変わらないものの, $z_x$が振動的になっていることがわかる.
  Noneについては本研究の制御を使った場合に比べて$E_z$が大きく, $x_z$が一度プラス方向に振れた後, 大きくマイナス方向にもう一度振れている.
  一方で本研究の制御を使った場合はマイナス方向にはあまり振れず, そのまま収束することがわかる.
  PDの場合はゲインを変更してもNoneよりも悪い方向になる場合がほとんどであった.

  次に, Proposed Collectionにおいて得られたモデルを使った場合について, PBを$(\theta_{s-p}, \theta^{offset}_{a-p})=\{(-5.0, -0.5), (5.0, 0.5)\}$の際に得られた値とした場合の結果を\figref{figure:sim-control2}に示す.
  どちらについても, $(\theta_{s-p}, \theta^{offset}_{a-p})=(0.0, 0.0)$の際に得られた正しいPBを使った場合に比べて大きく誤差がのってしまうことがわかる.

  最後に, Proposed Collectionにおいて得られたモデルを使った場合について, $C_{l}=30$は固定し, $(C_{f}, C_{u})=\{(0, 3), (1, 3), (0, 10)\}$のようにバランス制御のパラメータを変化させた際の結果を\figref{figure:sim-control3}に示す.
  なお, ここでは筋張力の前ステップからの時間変化のノルム$||\Delta\bm{f}||_{2}$の遷移についても表示しており, その値の二乗平均平方根を$E_f$とする.
  また, \figref{figure:sim-control3}の左上図は\figref{figure:sim-control1}のProposed Collectionと同じグラフである.
  $C_{u}$を3から10に変更すると, $\bm{u}=\theta^{ref}$の動きが抑制されるため, $z_{x}$の動きが\figref{figure:sim-control1}のNoneに近づくことがわかる.
  また, $C_{f}$を0から1に変更すると, $||\Delta\bm{f}||_{2}$のピークが収まり, $E_{f}$も0.674と$C_{f}=0$の場合に比べると小さな値となっていることがわかる.
}%

\begin{figure}[t]
  \centering
  \includegraphics[width=0.8\columnwidth]{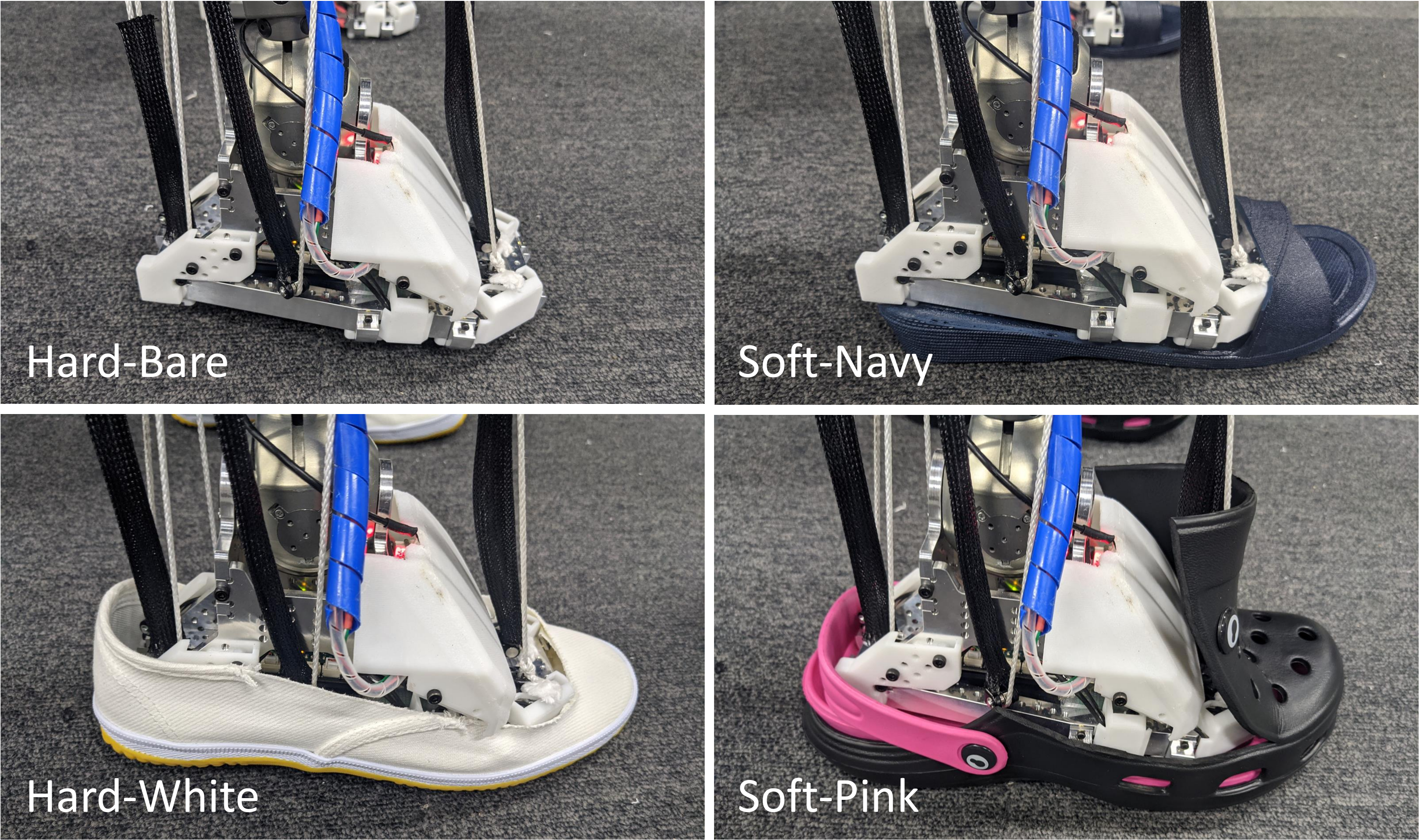}
  \vspace{-1.0ex}
  \caption{Various shoes used for the actual robot experiment as temporal body changes.}
  \label{figure:act-setup}
  \vspace{-3.0ex}
\end{figure}

\begin{figure}[t]
  \centering
  \includegraphics[width=0.99\columnwidth]{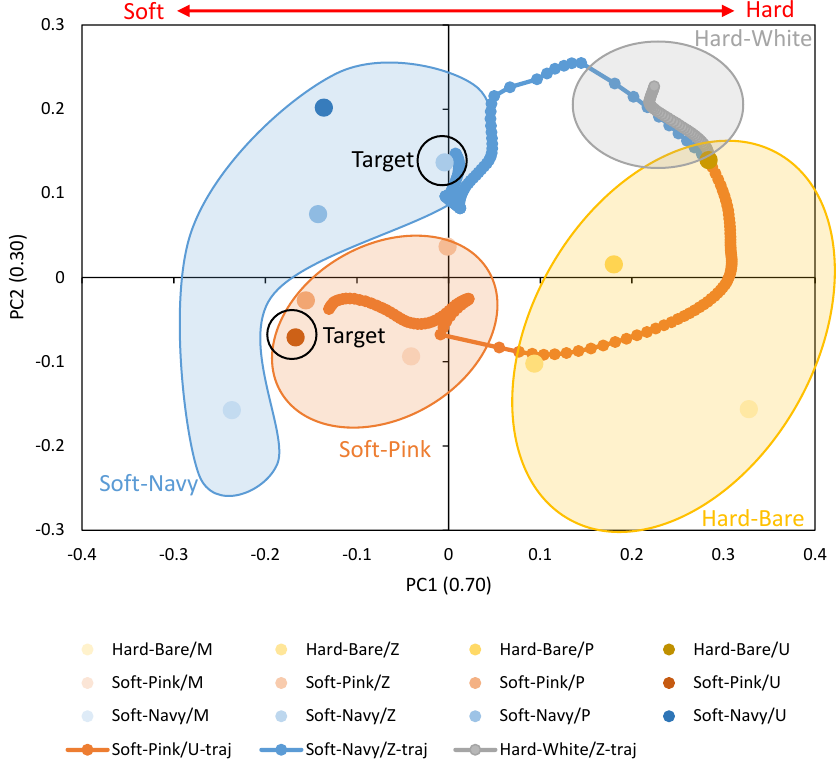}
  \vspace{-3.0ex}
  \caption{Actual robot experiment: the arrangement of parametric bias when training DPMPB using the data collected with Proposed Collection, and the trajectories of parametric bias when running online learning by setting the body state to Soft-Pink/U, Soft-Navy/Z, or Hard-White/Z.}
  \label{figure:act-pb}
  \vspace{-1.0ex}
\end{figure}

\begin{figure}[t]
  \centering
  \includegraphics[width=0.99\columnwidth]{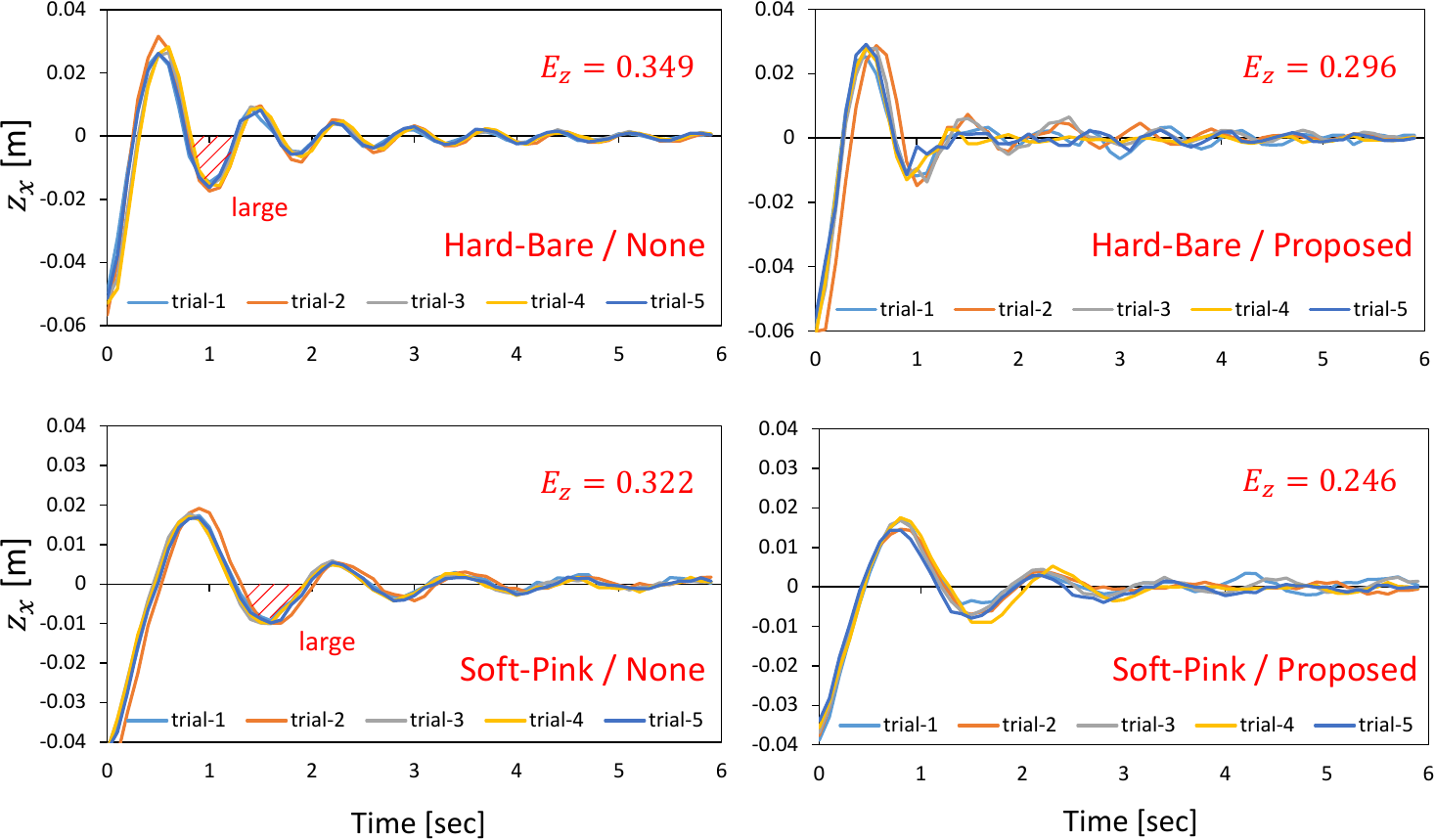}
  \vspace{-3.0ex}
  \caption{Actual robot experiment: the transitions of $z_x$ when 15 N (for Hard-Bare) or 10 N (for Soft-Pink) of external force is applied to the chest link and released, while the balance control using DPMPB is performed (Proposed), or while no control is performed (None).}
  \label{figure:act-control}
  \vspace{-1.0ex}
\end{figure}

\subsection{Actual Robot Experiment}
\switchlanguage%
{%
  \subsubsection{Training of DPMPB}
  In this experiment, we handle changes in the body state, such as which shoes to wear among \figref{figure:act-setup} \{Hard-Bare, Hard-White, Soft-Pink, Soft-Navy\}, and the posture of the upper body \{M, Z, P, U\} (M is at $\theta_{s-p}=-5$, Z is at $\theta_{s-p}=0$, P is at $\theta_{s-p}=5$, and U is at $\theta_{e-p}=-60$ [deg], where $\theta_{s-p}$ is the pitch angle of the spine joint and $\theta_{e-p}$ is the pitch angle of the elbow joint).
  First, we collect the data while changing the body state into 12 different types, by changing shoes to \{Hard-Bare, Soft-Pink, Soft-Navy\} and upper body posture to \{M, Z, P, U\} (referred to as Hard-Bare/U or Soft-Navy/Z).
  For each body state, we obtain data for 300 steps.
  Here, we only collect data by \equref{eq:collect}, and the trained balance control is denoted as Proposed, while the case without any control is denoted as None.
  Parametric bias obtained by training DPMPB with these data is plotted on a two-dimensional plane by applying PCA to it, as shown in \figref{figure:act-pb}.
  We can see that the space of PB is roughly structured for Soft-Navy, Soft-Pink, and Hard-Bare.
  It can also be seen that the upper body postures of P and U have similar dynamics in the sense that the robot leans forward, and that the PBs of P and U are relatively close to each other.
  For this model, fine tuning from DPMPB trained in simulation does not reduce the loss much because the dynamics is very different.

  \subsubsection{Online Update of Parametric Bias}
  We start with $\bm{p}$ in Hard-Bare/U and examine how $\bm{p}$ transitions when the online update of PB is executed at the same time as the body is moved as in the data collection.
  The trajectories of $\bm{p}$ when the current body states are Soft-Pink/U, Soft-Navy/Z, and Hard-White/Z are shown in \figref{figure:act-pb}.
  For Soft-Pink/U and Soft-Navy/Z, we can see that the current $\bm{p}$ gradually approaches the $\bm{p}$ trained in the same body state as the current one.
  Thus, it is possible to correctly recognize the body state by searching the space of $\bm{p}$.
  Although Hard-White is not included in the training data, it is placed near the upper part of Hard-Bare as a result of online learning.
  Shoes have various parameters such as shape, friction, and softness, but the soles of Hard-White and Hard-Bare, Soft-Pink and Soft-Navy are similar in hardness.

  \subsubsection{Balance Control Using DPMPB}
  In this experiment, the upper body posture is Z, and the transition of $z_x$ after applying a certain force to the waist link (15 N for Hard-Bare and 10 N for Soft-Pink) and then releasing it is examined five times for 6 seconds.
  The results of the balance control for Proposed and None are shown in \figref{figure:act-control}.
  For $z_x$, offsets are removed to align the origins of the plots, and the average of the sum of $|z_x|$ for 6 seconds (30 steps) is shown as $E_z$.
  For PB, we use the values obtained while training for each body state (Hard-Bare/Z or Soft-Pink/Z).
  For Hard-Bare/Z, we set $(C_{f}, C_{l}, C_{u})=(0, 30, 3)$, and for Soft-Pink/Z, $(C_{f}, C_{l}, C_{u})=(0, 3, 1)$.
  Although the effect is not as large as in the simulation, it can be seen that the convergence after the external force is faster in Proposed than in None.
  In fact, for Hard-Bare, $E_z=0.349$ for None and $E_z=0.296$ for Proposed, and for Soft-Pink, $E_z=0.322$ for None and $E_z=0.246$ for Proposed, indicating that Proposed has less error.
}%
{%
  筋骨格ヒューマノイドMusashiを使った実機実験を行う.
  ここでは身体状態変化として, \figref{figure:act-setup}のうちのどの靴を履くか\{Hard-Bare, Hard-White, Soft-Pink, Soft-Navy\}, また, 上半身の姿勢\{M, Z, P, U\}を扱う(ここで, Mは$\theta_{s-p}=-5$, Zは$\theta_{s-p}=0$, Pは$\theta_{s-p}=5$, Uは両腕の肘の関節角度$\theta_{e-p}=-60$ [deg]を表す).
  まずデータ収集を行うが, この際の身体状態について, 靴を\{Hard-Bare, Soft-Pink, Soft-Navy\}, 上半身姿勢を\{M, Z, P, U\}として12種類に変化させながらデータを取る(Hard-Bare/UやSoft-Navy/Zのように表記する場合がある).
  それぞれの身体状態について, 300ステップのデータを得る.
  ここでは\equref{eq:collect}によるデータ収集のみを行い, この方法で習得した制御をProposed, 何も制御しない場合をNoneと表記する.
  これらのデータを使ってDPMPBを学習した際に得られたParametric BiasにPCAをかけ, 2次元平面上にプロットしたものを\figref{figure:act-pb}に示す.
  シミュレーション実験ほど綺麗ではないが, Soft-Navy, Soft-Pink, Hard-Bareごとに大まかにPBが構造化されていることがわかる.
  また, 上半身姿勢のPとUは姿勢こそ違うものの前傾するという意味では似たダイナミクスを持っており, PとUは比較的近い位置にPB同士が存在することもわかる.
  本モデルについては, シミュレーションで得られたDPMPBからのFine Tuningではモデルが大きく異なるため損失があまり下がらなかった.

  次に, Parametric Biasのオンライン更新について実験を行う.
  $\bm{p}$をHard-Bare/Uの状態から始め, データ収集時と同様に体を動かすと同時にPBのオンライン更新を実行したときに, どのように$\bm{p}$が遷移するかを調べる.
  現在の身体状態をSoft-Pink/U, Soft-Navy/Z, Hard-White/Zとしたときの$\bm{p}$の軌跡を\figref{figure:act-pb}に示す.
  Soft-Pink/UとSoft-Navy/Zについては, 現在の$\bm{p}$が, 徐々に現在と同じ身体状態において学習された$\bm{p}$へと近づいていくことがわかる.
  つまり, $\bm{p}$の空間を探索して身体状態を正しく認識するような適応が可能である.
  また, Hard-Whiteについては学習時のデータには入っていないが, これは学習結果としてHard-Bareの近く上方あたりに配置された.
  なお, 靴には形や摩擦, 柔らかさなどの様々なパラメータがあるが, Hard-WhiteとHard-Bare, Soft-PinkとSoft-Navyはそれぞれ靴底の硬さが似通っている.

  最後に, 本研究の制御Proposedと何も制御しないNoneにおけるバランス制御の結果を\figref{figure:act-control}に示す.
  本実験では全て上半身の姿勢はZとし, 腰リンクに対してある一定の力(Hard-Bareについては15 N, Soft-Pinkについては10 N)をかけてからこれを離した後の$z_x$の遷移を6秒間, 5回ずつ検証する.
  $z_x$についてはグラフを揃えるためにオフセットを取り除き, 6秒間(30ステップ分)の$|z_x|$の合計値の平均を$E_z$として表示している.
  PBについてはそれぞれの身体状態(Hard-Bare/ZまたはSoft-Pink/Z)について学習時に得られた値を利用する.
  また, Hard-Bare/Zについては$(C_{f}, C_{l}, C_{u})=(0, 30, 3)$, Soft-Pink/Zについては$(C_{f}, C_{l}, C_{u})=(0, 3, 1)$とした.
  シミュレーションほど効果は大きくないが, 外力が加わった後の収束は, Noneに比べてProposedの方が速いことが見て取れる.
  実際, Hard-Bareについては, Noneでは$E_z=0.349$, Proposedでは$E_z=0.296$, また, Soft-Pinkについては, Noneでは$E_z=0.322$, Proposedでは$E_z=0.246$となり, Proposedの方が誤差が少ない.
}%

\section{Discussion} \label{sec:discussion}
\switchlanguage%
{%
  We discuss the experimental results of this study.
  First, the simulation results show that the parameters of the dynamics not explicitly given as values are embedded in parametric bias by learning the DPMPB.
  This arrangement of PB is self-organized nicely as the collected data has more diverse time-series changes, and PB can be updated online to adapt to the current dynamics.
  In addition, it can be seen that learning from data with various time series changes makes the balance control more accurate and the convergence of the response to external forces faster.
  In the case of no control or PID control, the convergence may be slow or divergence may occur, but our method enables the robot to stand upright stably and immediately after external force.
  On the other hand, when PB is not adapted to the current body state, the balance control may not work well due to the difference between the predicted dynamics and the actual dynamics.
  By changing the weights in the loss function, this balance control can simultaneously execute other objectives, such as reducing the control input and suppressing the changes in muscle length and tension.

  Second, in the actual robot experiment, we handled the difference in dynamics of shoes, which is difficult to be given as values explicitly by humans.
  The trained PBs are grouped according to the type of shoe, and it is possible to estimate the type of shoe that the robot is currently wearing based on the current motion and understand the dynamics.
  The space of PB is constructed to reflect the nearness and remoteness of the dynamics that could be generalized to shoes that are not used for training.
  In addition, upper body postures such as the elbow and hip angles can be treated in the same variable of PB, in the form of changes in the dynamics of the lower body.
  The balance control shows some performance, and the convergence of the error is faster than the case without the control.
  On the other hand, since it is difficult to align the experimental conditions in the actual robot, it is inevitable that the performance in the actual robot is lower than that in the simulation, and there is room for improvement in the future.

  The limitation of this study is described below.
  First, there is a problem that the speed of the iterative backpropagation becomes a rate-limiting factor and the balance control cannot be executed at a fast frequency.
  In this study, the limit is about 15 Hz, and the results are not much different from those of 5 Hz.
  It is found that if the period can be increased to about 100 Hz, the response to disturbances becomes faster, and the range of application will be expanded.
  On the other hand, the prediction accuracy of a trained model is likely an issue to be addressed in the future, since prediction errors accumulate and a long control horizon is required for high frequency.

  Second, there is a problem of data collection.
  In this study, we collected a variety of data by gradually shaking the body, but in order to obtain more dynamic data, we need to devise further ways of data collection, such as alternating between learning and data collection.
  If data collection becomes more efficient, it will be possible to handle not only simple balance control, but also more complex tasks such as stepping forward and walking, which require higher dimensional control inputs.
  In the future, it would be desirable to develop a curriculum learning method in which the robot learns to step while using a handrail, and gradually releases its hands when walking.

  We describe some future developments.
  It would be meaningful to practice scenarios in which the robot wears different shoes depending on the task, such as shoes that are easy to balance, shoes that allow fast movement, waterproof shoes, and so on.
  In addition, we would like to consider the environment as a part of the body, and work on walking considering changes in the ground, using assitive tools, etc.
}%
{%
  本研究の実験結果について考察する.
  まずシミュレーション結果から, DPMPBの学習により明示的に値として与えていないダイナミクスのパラメータがParametric Biasに埋め込まれることがわかった.
  このPBの配置はデータが多様な時系列変化を持つほど綺麗に自己組織化し, オンラインで現在のダイナミクスを学習しPBを正確に更新することができるようになる.
  また, 多様な時系列変化を学習することでよりバランス制御が正確になり, 外力に対する応答の収束が速くなることがわかる.
  制御をしない場合やPID制御をする場合には収束が遅かったり逆に発散してしまう場合があるが, 本手法により外力をいなし即座に安定して直立することが可能となった.
  一方で, PBが現在の身体状態に適合したものではない場合, 現実のダイナミクスと予測したダイナミクスが異なるためうまくバランス制御が働かない場合がある.
  本バランス制御は損失関数における重みを変化させることで, 制御入力を小さくしたり, 筋長や筋張力の変化を抑えたりと, 別の目的を同時に実行することが可能である.

  次に実機実験では, 靴という人間が明示的に与えることの難しいダイナミクスの違いを扱った.
  学習されたPBは靴の種類ごとにまとまりを示し, また現在の動きから現在履いている靴の種類を当て, そのダイナミクスを理解することができる.
  学習時に用いていない靴にも汎化するような ダイナミクスの近さや遠さが反映されたPBの空間が構築されていた.
  また, 肘や腰の角度のような上半身の姿勢についても, PBの中では下半身のダイナミクス変化という形で同じ変数で統一的に扱うことが可能であった.
  バランス制御についてはシミュレーションほどではないもののある程度の性能を示し, 制御を入れない場合よりも速い誤差の収束が可能であった.
  一方, 実機において実験条件を揃えることは難しく, 実機における性能がシミュレーションにおける性能よりも下がってしまうことは現状避けられず, 今後改善の余地がある.

  本研究のLimitationについて述べる.
  まず, 現状は誤差逆伝播の速度が律速となり, 速い周期で制御を実行できないという問題がある.
  本研究では15 Hz程度が限界であり, その程度では5 Hzと結果は大差なかった.
  100 Hz程度まで周期を上げることができれば外乱に対する応答はより速くなることが分かっており, 適用範囲も広がると考えられる.
  一方で, 長い制御ホライズンが必要であり予測誤差が蓄積してしまうため, 学習精度の向上も今後の課題となりそうである.

  次に, データ収集の問題が挙げられる.
  本研究では徐々に大きく体を揺らすことで多様なデータを収集したが, よりダイナミックなデータを得るためにはカリキュラム学習のように学習とデータ収集を交互に行うような, さらなるデータ収集の工夫が必要である.
  データ収集が効率的になれば, 単純なバランス制御だけでなく, 足を踏み出す, 歩くというより複雑で制御入力が高次元なタスクを扱えるようになると考える.
  今後は, 手すりを使いながら足踏みを学習し, 最終的に手を離して歩くようなカリキュラム型の学習方法の開発も望まれる.

  本研究の今後の発展について述べる.
  今回は様々な靴に対応するという実験を行っているが, 今後, バランスを取りやすい靴や速く動ける靴, 防水の靴等をタスクに応じて使い分けるというシナリオを実践することも有意義であると考える.
  また, 環境も身体の一部と考え, 地面の変化や道具を利用した歩行等にも取り組みたい.
}%

\section{CONCLUSION} \label{sec:conclusion}
\switchlanguage%
{%
  In this study, we proposed a deep predictive model learning method including parametric bias for balance control of complex musculoskeletal humanoids with flexibility and redundancy.
  For the task of balance control, it is difficult to collect data for in the actual robot.
  We can construct a stable balance control by collecting data while gradually increasing the random displacement of the control input and periodically changing its random width.
  In addition, the changes in upper body posture, the origin of joints and muscles, and footwear, which are not included in the dynamics model of the ankle, can be embedded as implicit changes in dynamics into parametric bias.
  Using the proposed DPMPB, the musculoskeletal humanoid successfully controls its balance according to various loss functions while adapting to changes in the body state.
  In the future, we would like to explore a method for autonomous learning of the foot-stepping motion using only the actual robot with assistance such as a handrail.
}%
{%
  本研究では柔軟性と冗長性を備える複雑な筋骨格ヒューマノイドのバランス制御に向けた, Parametric Biasを含む深層予測モデル学習の提案を行った.
  バランス制御というデータ収集の困難なタスクに対して, ランダムな制御入力の変位を徐々に増やし, かつそのランダム幅を周期的に変化させながらデータを収集することで, 安定した有用なバランス制御を構築することができる.
  また, 学習した足首に関するダイナミクスに含まれないような, 上半身の姿勢や履いている靴等については暗黙的なダイナミクス変化としてParametric Biasに埋め込むことができる.
  提案したDPMPBを使うことで, 身体状態の変化に適応しつつ, 様々な損失関数に従って筋骨格ヒューマノイドがバランス制御を行うことに成功した.
  今後は, 足を踏み出す動作を手すり等の補助を使いつつ, 実機だけで自律的に学習する手法について探索したい.
}%

{
  %\footnotesize
  %\small
  %\bibliographystyle{junsrt}
  \bibliographystyle{IEEEtran}
  \bibliography{main}
}

\end{document}